\documentclass[11pt]{article}

\usepackage{lscape}
\usepackage[margin=1in]{geometry}
\usepackage{cite}
\usepackage{amsmath,amssymb,amsfonts}
\usepackage{algorithmic}
\usepackage{graphicx}
\usepackage{textcomp}
\usepackage{xcolor}
\def\BibTeX{{\rm B\kern-.05em{\sc i\kern-.025em b}\kern-.08em
    T\kern-.1667em\lower.7ex\hbox{E}\kern-.125emX}}

\usepackage{balance}  
\usepackage{epsfig}
\usepackage{amssymb}
\usepackage{amsmath}
\usepackage{amsfonts}
\usepackage[lofdepth,lotdepth]{subfig}
\usepackage{times}
\usepackage{multirow}
\usepackage{multicol}
\usepackage{color}
\usepackage{capt-of}
\usepackage{caption}
\usepackage{marvosym}

\usepackage{algorithm}

\usepackage{scalerel}
\newcommand\reallywidehat[1]{\arraycolsep=0pt\relax%
	\begin{array}{c}
		\stretchto{
			\scaleto{
				\scalerel*[\widthof{\ensuremath{#1}}]{\kern-.5pt\bigwedge\kern-.5pt}
				{\rule[-\textheight/2]{1ex}{\textheight}} 
			}{\textheight} %
		}{0.5ex}\\           
		#1\\                 
		\rule{-1ex}{0ex}
	\end{array}
}

\begin{document}
	
\title{Are Outlier Detection Methods Resilient  to Sampling?
}

\author{Laure Berti-Equille$^1$ \and Ji Meng Loh$^2$ \and Saravanan Thirumuruganathan$^3$}
\date{$^1$UMR ESPACE-DEV, IRD-Maison de la T\'{e}l\'{e}d\'{e}tection, Montpellier, France\\%
$^2$Department of Mathematical Sciences, New Jersey Institute of Technology, New Jersey, USA \\%
$^3$Qatar Computing Research Institute, Hamad Bin Khalifa University, Doha, Qatar
}


\maketitle
\begin{abstract} 
        Outlier detection is a fundamental task in data mining and has many applications including detecting errors in databases. While there has been extensive prior work on methods for outlier detection, modern datasets often have sizes that are beyond the ability of commonly used methods to process the data within a reasonable time. To overcome this issue, outlier detection methods can be trained over samples of the full-sized dataset. However, it is not clear how a model trained on a sample compares with one trained on the entire dataset. In this paper, we  introduce the notion of {\it resilience  to sampling} for outlier detection methods. Orthogonal to  traditional performance metrics such as precision/recall, resilience represents the extent to which the outliers detected by a method applied to samples from a sampling scheme matches those when applied to the whole dataset. We propose a novel approach for estimating the resilience to sampling of both individual outlier methods and their ensembles. We performed an extensive experimental study on synthetic and real-world datasets where we study seven diverse and representative outlier detection methods, compare results obtained from samples versus those obtained from the whole datasets and evaluate the accuracy of our resilience estimates. We observed that the methods are not equally {\it resilient} to a given sampling scheme and it is often the case that careful joint selection of both the sampling scheme and the outlier detection method is necessary. It is our hope that the paper initiates research on designing outlier detection algorithms that are resilient to sampling.
      \end{abstract}%
	
%

	\sloppy
	
	\section{Introduction}
    Poor data quality is an important and challenging problem in data analysis.
    Outliers, which are data points that deviate from expected behavior are one of the four major categories of data quality errors 
(the others being duplicates, rule violations, and pattern violations).
outlier detection is a fundamental data analysis task with widespread applicability in a number of critical domains such as cybersecurity, fraud detection, etc. Not surprisingly, there has been extensive prior work (see Section~\ref{sec:relWork} for summary) on identifying outliers under a wide variety of scenarios. However, many of these approaches do not scale well to large datasets that are now common. 

Parametric methods assume an underlying data distribution (e.g.\ Gaussian mixture) $M$, estimate the parameters from the dataset and identify a given point $t$ as an outlier if $P(t|M)$ is smaller than a given threshold. Estimating these parameters on the entire dataset can be computationally expensive. Non-parametric methods using distance-based techniques often do not require distributional assumptions, and a data point is categorized as an outlier if it lies far away from other data points. Such approaches have scalability issues since they might require computation of all pairwise distances. 
Adaptations such as indexes, approximations and pruning for distance computations only partially address scalability issues.

This problem cannot be solved by choosing the most scalable algorithm. The particular algorithm might not have the best detection rate or it might not be the most suitable for the problem at hand. 
Furthermore, in outlier ensembles that combine multiple detection methods, even one non-scalable constituent algorithm affects the entire ensemble.

	\begin{figure*}[!t]
		\centering
		\includegraphics[scale=0.25]{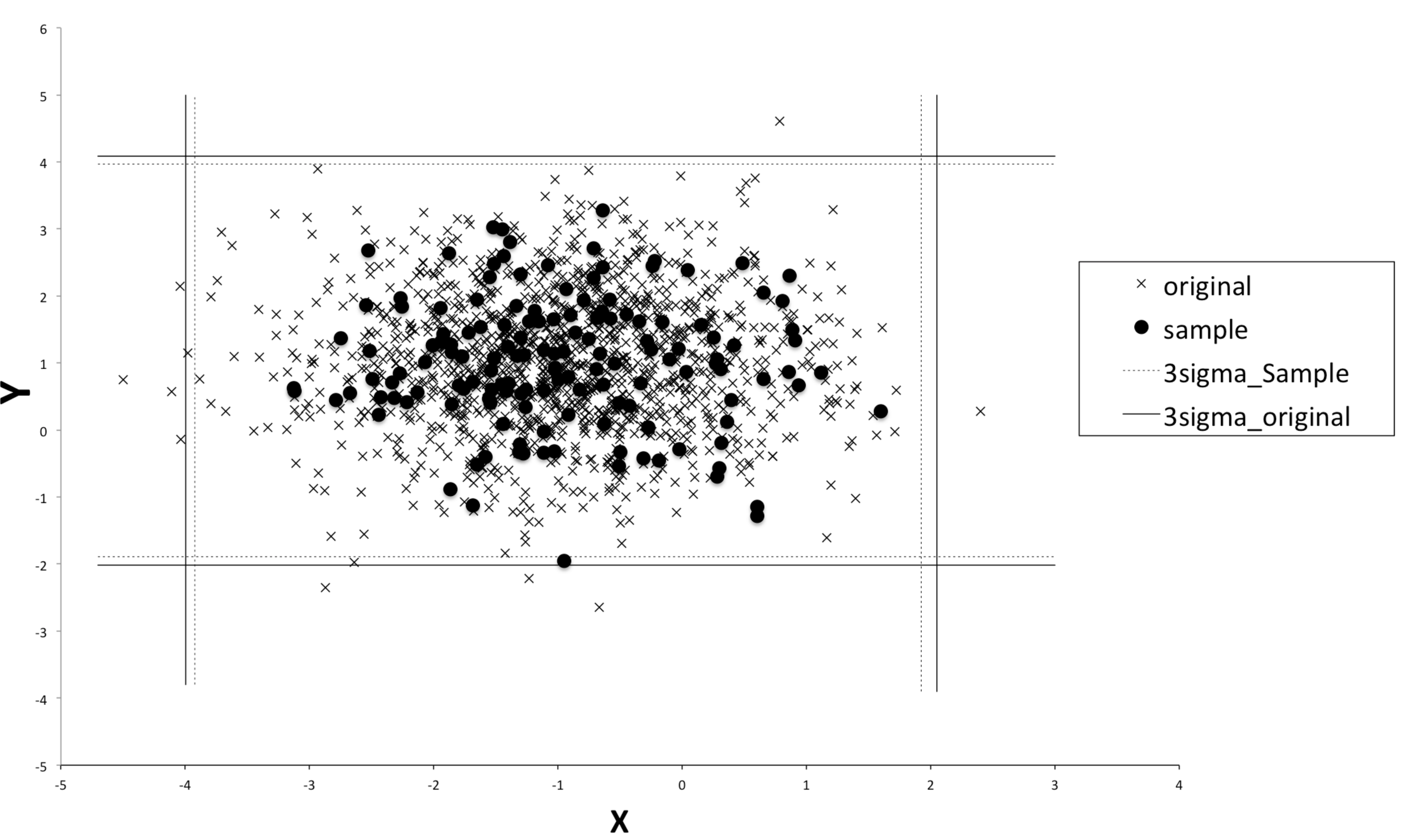}
		\includegraphics[scale=0.25]{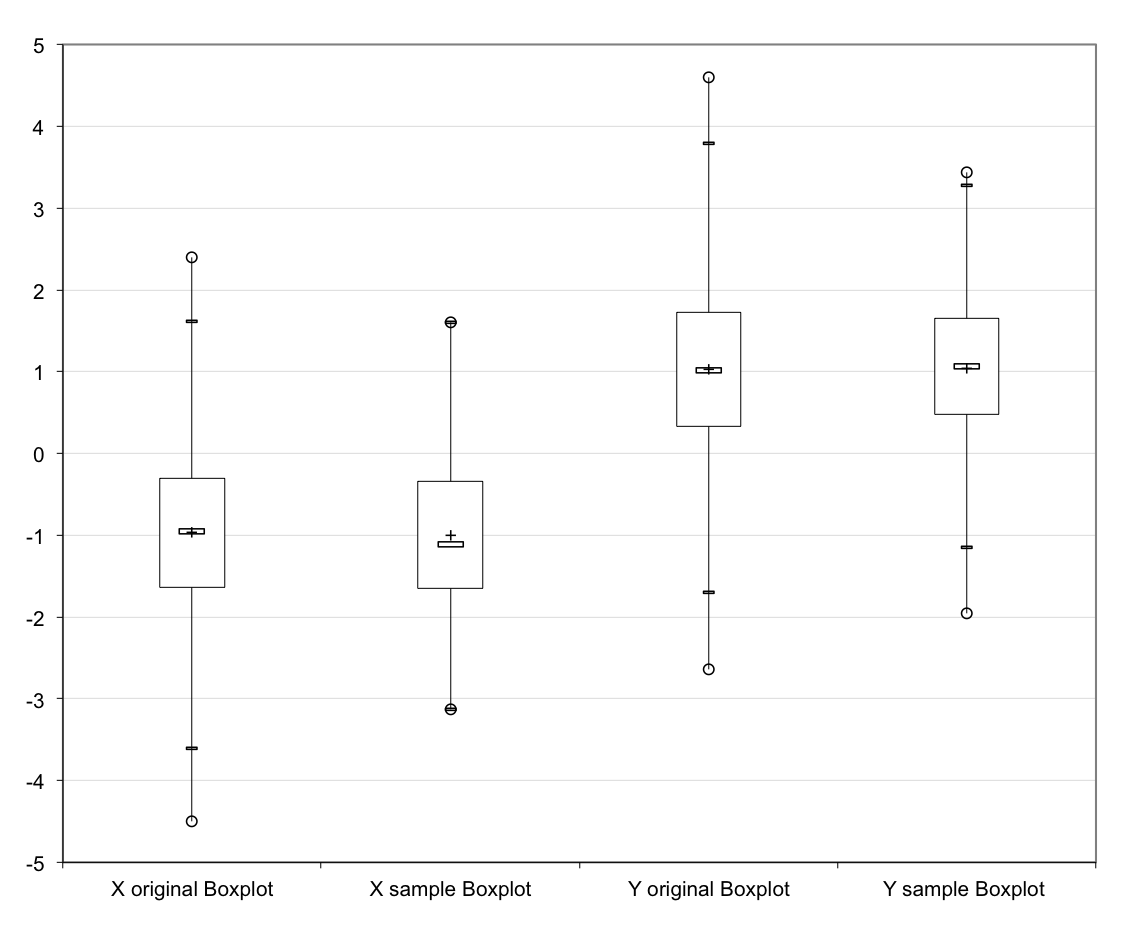}
		\caption{Generated data ($\times$) and a 10\% random sample ($\bullet$). We applied $3\sigma$ (left) and boxplot (right) methods. The two methods have different behaviors in terms of resilience: boxplot results are not preserved when sampling, while the $3\sigma$ limits for the original dataset (solid lines) and for the sample (dashed lines) are very similar.}\label{fig:monovariate}
	\end{figure*}


{\bf Is Sampling the Silver Bullet?} A seemingly viable solution is to build models for outlier detection over samples of the entire dataset. While sampling reduces the amount of computations required for parameter estimation and pairwise distance calculations, it is not the panacea to our problems. This is because the results of detection methods on samples and on the entire dataset may not agree.

We illustrate this with an empirical example. We generated a bivariate distribution of 1,500 data points with mean $(-1,1)$ and variance $(1.015, 1.035)$, plotted  ($\times$) in Figure \ref{fig:monovariate} (left), and extracted a 10\% random sample ($\bullet$). We used two {\em univariate} outlier detection methods on both sample and entire dataset.

The first univariate method, dubbed $3\sigma$, identifies all observations outside the interval between mean $\mu$ and $\pm 3\sigma$ as outliers. The second method uses the boxplot inner $(Q1-1.5\times IQR)$ and outer $(Q3+1.5\times IQR)$ fences, $Q1$ and $Q3$ being the first and third quartiles of the data and $IQR = Q3 - Q1$. Points beyond the inner and outer fences are deemed outliers. Figure~\ref{fig:monovariate} shows that the $3\sigma$ approach performs similarly with both sample and dataset while the boxplot method does not. Informally, this is because the sample mean and variance are better estimates of the population mean and variance than sample quartiles are of the population counterparts. We also found similar divergent behavior with two {\em multivariate} detection methods, $\chi^2$ and LOF (not shown).
	
	
{\bf Outline of Technical Contributions:} In this paper, we conduct a systematic study of how sampling impacts outlier detection algorithms. The paper has four contributions.
	
	\noindent{\bf (1) Formalization of Resilience:}
	The examples  above show that while some outlier detection methods generalize well from samples to the entire dataset, others do not. We introduce and formalize the notion of {\em resilience to sampling} of outlier detection methods. Informally, resilience represents the extent to which the number and identification of detected outliers by a method applied to the whole dataset are preserved if it is actually applied to a sample. We emphasize that this property is orthogonal to traditional performance metrics such as specificity and sensitivity. We believe that understanding the notion of resilience is of paramount importance for practitioners. In many real-world scenarios, building outlier detection models over entire datasets is simply infeasible. Instead they are often built over samples where it is preferable to choose highly resilient methods over ones with lower resilience.
	
	\noindent{\bf (2) Estimation of Resilience:} 
	If the ground truth is available, estimating the resilience of an outlier detection method is straightforward. For the more realistic case of no ground truth, we propose a novel method to estimate the resilience. This involves using multiple samples from the entire dataset to build outlier models and perform differential analysis.
	
\noindent{\bf (3) Resilience and Outlier Ensembles:}
If an algorithm is resilient to sampling, then an effective model can be built over the sample instead of the entire dataset. Such efficiency gains allows the use of more sophisticated models such as ensembles: a set of resilient detection algorithms can be combined effectively as outlier ensembles. Since detection methods capture different characteristics of the data, outlier ensembles can leverage the combined judgment of the different methods to produce a consensus.
While the weights of an ensemble model is easily determined in traditional classification, outlier ensembling without ground truth is not straightforward. Is it possible to infer the ``ground truth'' using some mild assumptions and the output of individual outlier detection methods? We investigate how the key ideas of resilience can be used to build outlier ensembles and adapt the classical Dawid-Skeene's EM-based approach to obtain a weighted majority.
	
	\noindent{\bf (4) Extensive and Comprehensive Experiments:} We conducted an extensive study of resilience for seven diverse and representative outlier detection methods under two sampling schemes (uniform, block) and a subset-based partitioning approach over real-world and synthetic datasets. Our main finding is that the methods are not equally resilient to a given sampling scheme and hence it is important to 
understand the impact of the joint selection of sampling scheme and outlier detection method.

\section{Notation and data model}
\label{sec:overview}

Let the dataset $\mathbf{D}$ be an $N\times V$ matrix with $N$ records and $V$ variables, with each row $\mathbf{D}_i \sim f_1 + \gamma f_2$, a mixture of two distributions, the regular distribution $f_1$ and the outlier distribution $f_2$, independently for $i=1, \ldots N$. The parameter $\gamma$ is the probability of any data point being an outlier, or equivalently, the proportion of outliers. The value of attribute $j$ for record $i$, noted $x_{ij}$ has an associated detection result vector $\mathbf{o}_{ij}$ with $M$ elements corresponding to the detection results of  $M$ detection methods. If method $m$ identifies $x_{ij}$ as an outlier, then $\mathbf{o}_{ij}(m) \equiv o_{ijm} = 1$. 
 We note $O_m$ the set of outliers detected by method $m$ with size $|O_m|$.
\begin{table}[t]
\centering
\begin{tabular}{|l|p{7cm}|}
	\hline
	\multicolumn{2}{|c|}{ {\bf Notations}  } \\ 
		\hline
	$\mathbf{D}$ &  the whole dataset \\
	\hline$S$ &  the dataset sample of $\mathbf{D}$ \\
	\hline $N$ & the number of records of 	$\mathbf{D}$\\ 
	\hline $V$ & the number of variables of	$\mathbf{D}$\\
	\hline $M$ & the number of outlier detection methods\\
	\hline  $x_{ij}$& the value of  attribute $j$ for record $i$\\
	\hline $O_m$ & the set of outliers detected by method $m$ on $\mathbf{D}$ \\ 
	 \hline $O_m^S$ & the set of outliers detected by method $m$ on sample $S$ \\ 
    \hline $O_m[S]$ & the subset of outliers detected by method $m$ on the whole dataset filtered to contain only the sampled records \\
	\hline 
\end{tabular}
\caption{Notations}\label{tab:notations}

\end{table}

A sample $S$ has size $|S|$, consisting of $|S|$ rows of $\mathbf{D}$ and, in the context of massive datasets, usually we have $|S|\ll N$, so the sample can be loaded into the computer's memory. $O^S$ is the set of outliers detected from sample $S$ and $O[S]$ the  subset of outliers detected from the whole dataset that are part of the sample $S$. Table \ref{tab:notations} summarizes our notation. We are interested in comparing $O^S_m$ and $O_m[S]$ for each method $m$. We use $\alpha_m$ for the probability that method $m$ detects a true outlier in the whole dataset, also known as its {\em sensitivity},  {\em recall} or {\em true positive rate (TPR)}: $\alpha_m=\big[\frac{TP}{TP+FN}\big]_m$ with {\it TP} being the proportion of true positives and {\it FN}, the proportion of false negatives. When applied to a sample $S$, this probability is denoted $\alpha_m^S$. We use $\beta_m$ to denote the {\em specificity}, the probability that method $m$  correctly identifies negatives in the full dataset (and $\beta_m^S$ in the sample $S$), i.e.\ $\beta_m=\big[\frac{TN}{FP+TN}\big]_m$ with  {\it TN} the proportion of true negatives and {\it FP} the proportion of false positives.

\section{Resilience formalization} \label{sec:SingleAlgoSingleSample}
A method is said to be resilient to a sampling strategy if, when applied to the samples, it detects the same outliers as the ones the method would have detected if it was applied to the original whole dataset. 
We consider resilience as an orthogonal notion to the traditional quality metrics of precision and recall. 

Precision and recall are defined in an {\bf absolute} sense with respect to the ground truth. If the ground truth is not available, precision and recall cannot be computed. On the other hand, resilience is defined in a {\bf relative} sense as to what would have obtained if the method was applied to the whole dataset. 

Notice that it is not necessary to have ground truth to estimate resilience, suggesting that the precision and recall of a method could have an orthogonal impact on its resilience. Hence, a method can have good recall and precision but low resilience to a given sampling scheme, e.g.\ if the model built over the sample is a poor approximation of the one built over the entire dataset resulting in divergence of the outliers detected. Conversely, a method can be resilient but have low precision and recall. This might occur if the data characteristic used by the method is not discriminative of the full outlier distribution, but is well captured well in the sample.

Resilience measures similarity of performance of an outlier detection method applied to samples to that applied to the whole dataset. A straightforward method to capture this property is to redefine the notion of precision and recall such that ``ground truth'' corresponds to whether a data point is categorized as an outlier by the method trained over the entire dataset. While the redefined metrics can be used directly, it is more meaningful to express this property as a single metric. The concept of F1-measure combines precision and recall as their harmonic mean and
we define the resilience as the harmonic mean of the redefined precision and recall metrics.

\noindent{\bf Definition 1. Resilience to sampling. } 
Given a method $m$ and a sample $D_S$ of the full-sized dataset $\mathbf{D}$, the resilience, denoted $\rho_m(D_S)$ is defined as
\begin{eqnarray}\label{rho}
\rho_m(D_S)&=& \frac{2|O^S \cap O[S]|}{|O^S| + |O[S]|} \\ 
&=& \frac{2|O^S \cap O[S]|}{2|O^S \cap O[S]|+ |O^S \setminus O[S]|+ |O[S] \setminus O^S |}. \nonumber 
\end{eqnarray}

	\begin{figure}[!t]
	\centering
	\includegraphics[width=0.75\linewidth]{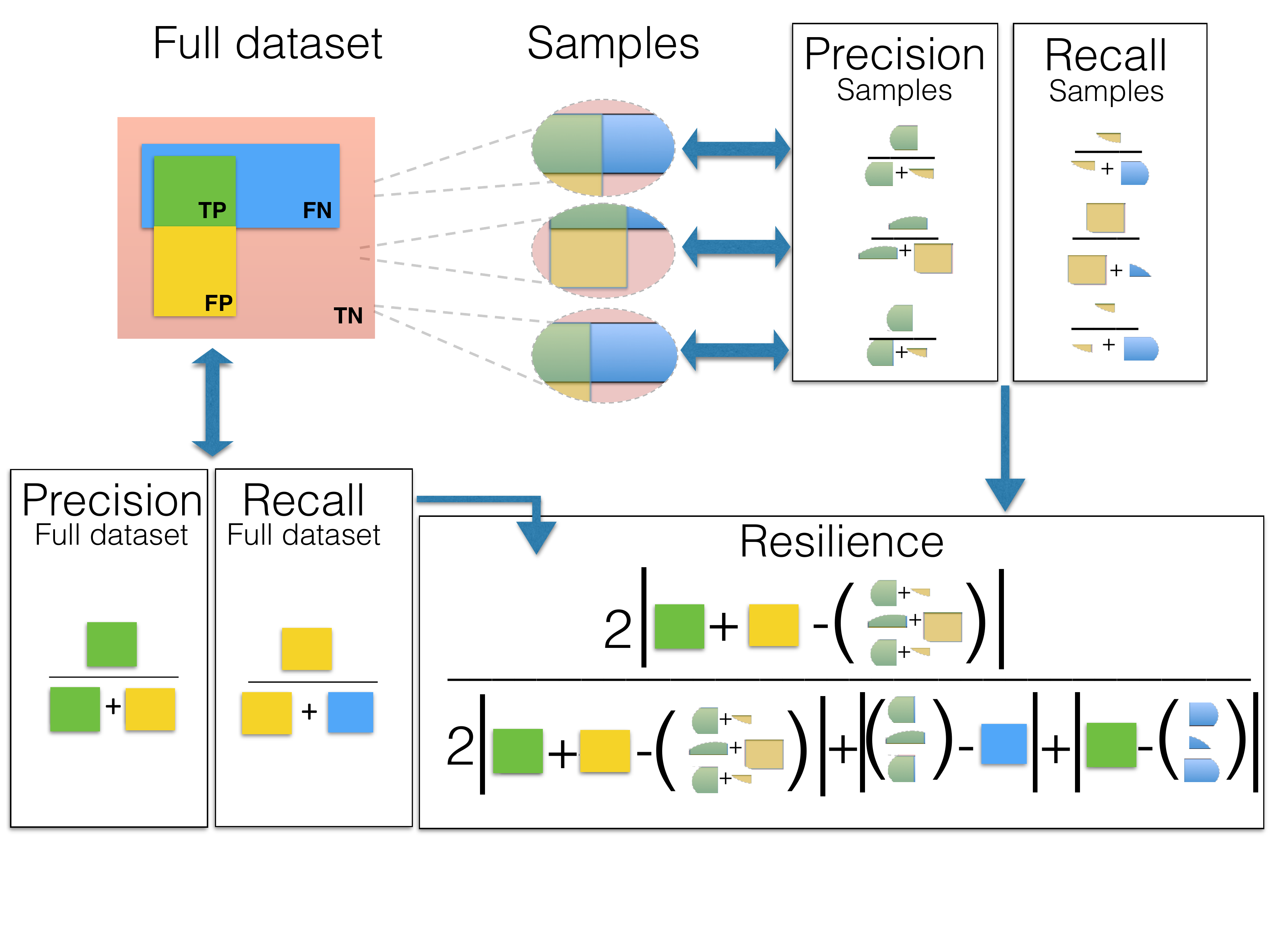}
	\caption{Relationships between precision, recall, and resilience }\label{fig:resilience}
\end{figure}

Figure \ref{fig:resilience} illustrates the relationships between resilience and precision/recall when the method is applied to the full dataset and also when applied to samples, Resilience is a combination of two sets of metrics, on $TP$, $FP$, and $FN$ from both the whole dataset and the samples. Similarly, the sensitivity (or recall) and specificity (or true negative rate) of a method can be computed either from the sample or the whole dataset, when ground truth is available and the resulting sets of detected outliers can be quite different. The resilience definition precisely quantifies the extent to which a method's results (good or bad) are preserved through sampling. 


When there is no ground truth,  we express the expected values of the quantities in Eq. (\ref{rho}) in terms of the sensitivity and specificity. Here we omit the index of the method $m$ for notation simplicity. The expected number of records that are detected in both the whole dataset and the sample is
\begin{equation}\label{WP-hat}
\begin{aligned}
E(|O^S \cap O[S]|)&=|S|\left[ \gamma\alpha \alpha^S + (1-\gamma)(1-\beta)(1-\beta^S)\right],
\end{aligned}
\end{equation}
where the terms within the square brackets on the right-hand side refer respectively to records correctly detected as outliers both in the whole dataset and in the sample, and to records incorrectly detected as outliers both in the whole dataset and in the sample. They involve the sensitivities $\alpha$ and $\alpha^S$ of the method when applied to the whole dataset and to the sample respectively, as well as the corresponding specificities $\beta$ and $\beta^S$. Similarly, we have
\begin{equation}\label{noWN-hat}
\begin{aligned}
\mbox{E}(|\overline{O^S} \cap \overline{O[S]}|)
& =	 |S|\left[(1-\gamma)\beta\beta^S + \gamma(1-\alpha)(1-\alpha^S) \right]. 
\end{aligned}
\end{equation}
The terms on the right-hand side refer to records corrected or incorrectly detected as inliers both in the whole dataset and in the sample. Lastly, we have
\begin{equation}\label{noWP-hat}
\begin{aligned}
\mbox{E}(|O^S \cap \overline{O[S]}|)& =|S| \left[\gamma(1-\alpha)\alpha^S + (1-\gamma)\beta(1-\beta^S) \right],\\
\mbox{E}(|\overline{O^S} \cap O[S]|) & =|S| \left[\gamma\alpha(1-\alpha^S) + (1-\gamma)(1-\beta)\beta^S \right].
\end{aligned}
\end{equation}

If detection results from the whole dataset are available, the resilience can be trivially computed by comparing with the sample detections. In the more realistic scenario where this is not available, the resilience has to be estimated using only the sample detections.
In Section \ref{sec:MultipleAlgosSingleSample}, we present a procedure to estimate the resilience when the whole dataset detection is unavailable, focusing on estimating the resilience of outlier ensembles. We estimate the sensitivities and specificities, then use them in Eqs.\ (\ref{WP-hat}), (\ref{noWN-hat}), and (\ref{noWP-hat}), to finally estimate the resilience with Eq.\ (\ref{rho}).

\section{Resilience for outlier ensembles} 
\label{sec:MultipleAlgosSingleSample}

The ability to build resilient models over small samples provides a number of advantages.
The computational benefits  accrued by running on a small sample for model building 
enables a user to invoke multiple outlier detection algorithms simultaneously. 
The information gained can be richer than invoking a single outlier detection algorithm over the entire data. Different outlier detection algorithms are often customized for different characteristics of datasets and outliers. Hence, it is plausible that by incorporating the results of multiple algorithms, one can achieve better outlier detection performance.
In this section, we make two contributions. 
First, we propose a mechanism  to ``ensemble'' multiple detection algorithms where each is invoked on a single sample. 
Second, we extend the notion of resilience for an outlier ensemble.

Ensemble analysis of outlier detection has received increasing attention from the research community. A common technique in building classifier ensembles is to combine the predictions of multiple classifiers using a weighted majority (or sum) rule. Classifiers that are more accurate are often provided with higher weights. However, adapting such a technique to outlier ensembles is much trickier since outlier detection is often an unsupervised learning problem with no ground truth. If we know the consensus of the outlier ensembles (say through weighted majority), we can easily compute the weight of an individual outlier detection algorithm based on its accuracy. However, to know the consensus of the outlier ensembles, we need the weight of each algorithm!
This conundrum is not specific to outlier ensembles and has been observed in other fields.
This chicken and egg issue is typically solved through iterative techniques such as EM (Expectation Maximization) that {\em jointly} estimate both the error rates and ground truth.  

{\bf Two Coin Model for Outlier Ensembles.} 
We use the Dawid-Skene model \cite{dawid1979maximum} to obtain ground truth 
using noisy labels from individual outlier detection algorithms.
In Section~\ref{sec:overview} we described the performance of an outlier detection algorithm in terms of its sensitivity and specificity relative to an unknown ground truth.
We use a simple generative process by which each outlier detection algorithm determines if a data point is an outlier or an inlier.
Consider a data point $x$ with true label $y$. Suppose the outlier detection method $m_j$ produces output $y_j$ (assumed to be boolean, outlier or inlier, for simplicity) through the following stochastic process:
If $y=1$, $m_j$ flips a coin with bias $\alpha_j$ (representing its sensitivity).
If $y=0$, $m_j$ flips a coin with bias $\beta_j$ (its specificity).
In both cases, $m_j$ returns $y$ if the tossed coin returns heads and the alternate label if the coin returns tails.
This generative model is very simple and only depends on $y$.
However, we note that there are extensions of \cite{dawid1979maximum} that can take into account other relevant factors including the observation $x$.

{\bf Jointly Learning Sensitivity, Specificity and Ground Truth.} 
To specify the algorithm, we first introduce the following notations.
Let $p_o$ and $p_i$ be the probability that a data point is an outlier and inlier respectively.
Ideally, $p_o = \gamma$, the outlier prevalence rate defined in Section \ref{sec:overview}. However, since the ground truth is unavailable, this value has to be estimated.
Each method has an associated confusion matrix, the probability that it outputs a particular label (say $b$) when the true label is $a$, denoted $\pi_{ab}^{m}$, where $a$ and $b$ can take values $o$ and $i$ (for outliers and inliers respectively).  
Thus, $\pi_{io}^m$ and $\pi_{oi}^m$ are the probabilities that method $m$ mistakes an inlier for an outlier and vice-versa
while $\pi_{oo}$ and $\pi_{ii}$ are the probabilities that it will identify the outlier/inlier correctly. 
Let $T_{ij}$ be an indicator variable equal to $1$ if $j$ is the true label for data point $i$ and $0$ otherwise, while $o_{ijm}$ be an indicator equal to $1$ if method $m$ assigned label $j$ to data point $i$.

We use an EM-like approach that iteratively performs the following:
(1) Estimate the outlier ground truth from different detectors taking into account their sensitivity and specificity; (2) Estimate the sensitivity and specificity of the detectors by comparing their results to the {\em inferred} ground truth.
Since the ground truth is unavailable, we treat it as a latent variable. Note that $\pi_{ab}^m$ can be computed as the fraction of data points for which method $m$ 
produces class label $b$ when the truth is $a$, to the total number of data points where $a$ is the true label. The prior probabilities $p_a$ can be computed as the fraction of the dataset that is assigned the label $a$ after ensembling. See Algorithm~\ref{alg:dawidSkene} below and \cite{dawid1979maximum} for additional details. 

\begin{equation}
    \label{eq:dsEstimatePi}
    \pi_{ab}^m =  \frac{ \sum_{i=1}^{|S|} T_{ia} \times o_{ibm}}{ \sum_{i=1}^{|S|} T_{ia}}  \quad \text{with } a, b \in \{\mbox{outlier}, \mbox{inlier}\}
\end{equation}
\begin{equation}
    \label{eq:dsEstimateP}
    p_j = \frac{\sum_{i=1}^{|S|} T_{ij}}{|S|} \quad \mbox{where } j \in \{\mbox{outlier}, \mbox{inlier}\}
\end{equation}
\begin{equation}
    \label{eq:dsEstimateGT}
    p(T_{ij} = 1 | S) \propto  \prod_{k=1}^{M} \prod_{\forall l \in \{\mbox{\footnotesize outlier}, \mbox{\footnotesize inlier}\}} \pi_{jl}^{k} \times p_j
\end{equation}

\begin{algorithm}
    \caption{Outlier Ensembling Algorithm}
    \label{alg:dawidSkene}
    \begin{algorithmic}[1]
        \STATE Initialize error rates $\pi^m_{a,b}$ for each method $m \in [1, m]$ and $a, b \in \{outlier, inlier\}$ (say to 0.5)
        \STATE Initialize prior probabilities $p_i = p_o = 0.5$
        \WHILE {estimated ground truth labels does not change between iterations }
            \STATE Estimate label $L_i$ for each record $D_i$ using Eq.~(\ref{eq:dsEstimateGT})
            \STATE Estimate error rates $\pi^m_{a,b}$ for each method $m$ using $L_i$ and $o_{iam}$ through Eq.~(\ref{eq:dsEstimatePi})
            \STATE Estimate class prior probabilities using $L_i$ and the $o_{iam}$ through Eq.~(\ref{eq:dsEstimateP})
        \ENDWHILE
        \RETURN error rates $\pi^m_{a,b}$ for each method and estimated labels $L_i$ for each record
    \end{algorithmic}
\end{algorithm}

Our proposed approach sidesteps a number of issues of prior outlier ensemble approaches such as outlier score normalization. By measuring the algorithm's error rates (through sensitivity and specificity), we can correct its bias and recover the ground truth with higher quality.
For example, detection algorithms that are conservative by rating more inliers as outliers can be easily identified and their bias corrected.

{\bf Resilience for Outlier Ensembles.} 
Given an outlier ensemble $E_O$ and a sample $S$, we can readily extend the notion of resilience to ensembles. Our definition is orthogonal to the specific ensembling approach, so that it applies to outlier ensembles other than our EM-based algorithm for ensembling (see Section~\ref{sec:relWork} for other ensembling approaches).  
Intuitively, we treat the ensemble as a complex black-box outlier detection algorithm and reuse the prior definition of resilience. Specifically, resilience for an outlier ensemble represents the extent to which the identification of outliers detected by the 
ensemble is preserved between component models that were trained on the entire dataset versus single sample respectively.
Specifically, if the detection results from the whole dataset are available, we can use Eq.~(\ref{rho}) to compute the resilience by comparing outlier detection results for whole dataset versus the sample. Without this information, we instead estimate the sensitivity and specificity of the outlier
ensemble using Equations~(\ref{WP-hat}), (\ref{noWN-hat}), and (\ref{noWP-hat}) and then plug them into Equation~(\ref{rho}).

\section{Experiments and discussion}
\label{sec:discussion}
\subsection{Experimental Setting}
\begin{figure}[!t]
	\centering
	\includegraphics[width=.5\linewidth]{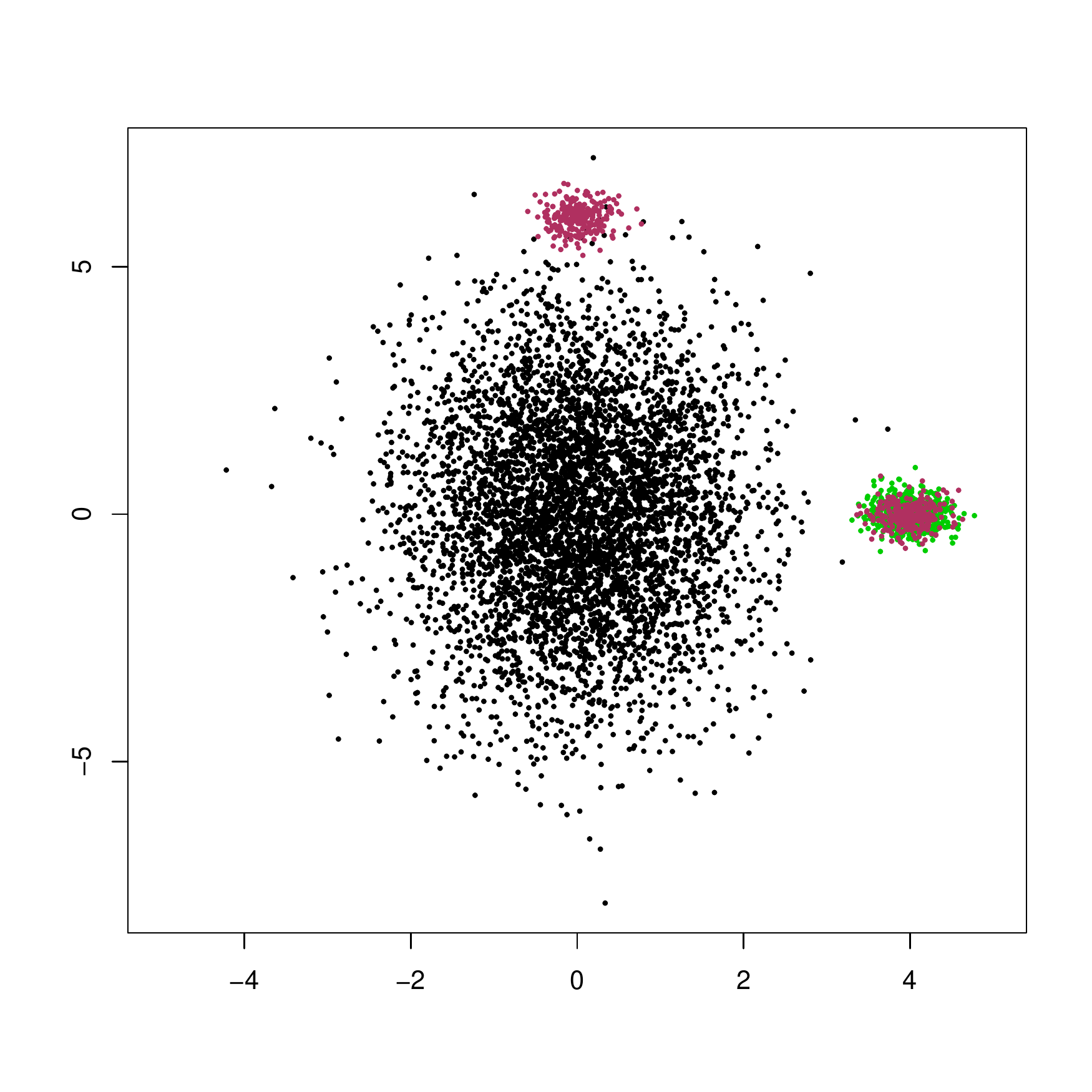}	
	\caption{Synthetic independent bivariate normal data (black) with outliers
injected at rates 1\%, 5\%,  and 10\% from the $N((4,0);.25)$ (green) and $N((0,6);.25)$ (maroon).}
	\label{fig:syn-data}
\end{figure}

\noindent{\bf Datasets.} We set up extensive experiments with eight real-world datasets and their variants from UCI\footnote{\scriptsize{{https://archive.ics.uci.edu/ml/datasets.html}}}, ODDS\footnote{\scriptsize{{http://odds.cs.stonybrook.edu/}}}, and  \cite{DAMI16}\footnote{\scriptsize{{http://www.dbs.ifi.lmu.de/research/outlier-evaluation/}}}. Their characteristics are shown in Table \ref{tab:datasets} and Figure
\ref{fig:syn-data} shows the synthetic data. The eight real datasets are diverse in terms of number of attributes, outlier injection rate, etc and often used in outlier detection research. We also used variants of the real datasets with and without ground truth ($GT$) outlier labels. The variants have different degrees of preprocessing which may have an impact on outlier detection performance. Synthetic datasets with 1,000, 5,000, and 10,000 records were generated from an independent bivariate normal with mean $(0,0)$ and standard deviations $1$ and $2$. Outliers were randomly inserted from two  outlier distributions. One is an independent bivariate normal distribution with mean $(4,0)$ and equal standard deviations $0.25$ (Fig. \ref{fig:syn-data}, green), and the other has an additional independent bivariate normal component with mean $(0,6)$ and the same standard deviation (maroon). These are injected at three different rates (1\%, 5\%, and 10\%).  We averaged results over 100 replications.

\begin{landscape}
\begin{table*}[!t]
	\centering
	\caption{Characteristics of the real-world dataset from UCI$^1$, ODDS$^2$ and [5]$^3$, and synthetic datasets used in the experiments ($N=$ number of observations, $V=$ number of variables, $GT=$ percentage of ground truth outliers, if available, $S=$ sample size, and $psize=$ percentage of sample size to whole dataset size).}
	\small
	\begin{tabular}{|p{1.7cm}|r|r|r||c|c|c|}
		\hline
		\multicolumn{1}{|c|}{{\bf Dataset name}}&\multicolumn{3}{|c||}{{\bf Dimensions} } & \multicolumn{3}{c|}{{\bf Sampling size $S$ }}	\\
		\cline{2-7}
		\multicolumn{1}{|c|}{{}}	&	\multicolumn{1}{|c|}{$N$} &	\multicolumn{1}{|c|}{$V$} &\multicolumn{1}{|c||}{$GT$}
		& Random ($psize$) & \multicolumn{1}{c|}{ Blocking (number of blocks -- block size) } &Subsets\# \\
		\hline
		Arrhythmia$^1$&452&258&\multicolumn{1}{|c||}{-}&20 (4.425\%); 50 (11.062 \%)&(1-20) (5-4) (10-2) (1-50) (5-10) (10-5)&\\
		Arrhythmia$^3$&452&274& 66 (15\%) &20 (4.425\%); 50 (11.062 \%)&(1-20) (5-4) (10-2) (1-50) (5-10) (10-5)&\\
		\cline{1-6}
		Diabetes$^1$	&768&9&\multicolumn{1}{|c||}{-}&50 (6.51 \%); 100 (13.021 \%)&(10-5) (25-2) (50-1) (10-10) (25-4), (50-2)&\\
		Diabetes$^3$ &768&8&268 (35\%)&50 (6.51 \%); 100 (13.021 \%)&(10-5) (25-2) (50-1) (10-10) (25-4), (50-2)&\\
		\cline{1-6}
		Waveform$^1$&5,000&22&\multicolumn{1}{|c||}{-}&250 (5\%);  500 (10\%)&(10-25)(25-10)(50-5)(10-50)(25-20)(50-10)&\\
		Waveform$^2$&3,443&21&100 (29\%)&250 (7.261\%); 500 (14.522\%)&(10-25)(25-10)(50-5)(10-50)(25-20)(50-10)&\\
		\cline{1-6}
		Spambase$^1$&4,601&58&\multicolumn{1}{|c||}{-}&200 (4.346\%); 400 (8.693\%)&(10-20) (25-8) (50-4) (10-40) (25-16) (50-8)&5; 10; 20\\
		Spam$\_02^3$&2,479&57&51 (2\%)&200 (8.068\%); 400 (13.136\%)&(10-20) (25-8) (50-4) (10-40) (25-16) (50-8)&\\
		Spam$\_05^3$&2,661&57&133 (5\%)&200 (7.516\%); 400 (15.032\%)&(10-20) (25-8) (50-4) (10-40) (25-16) (50-8)&\\
		Spam$\_10^3$&2,808&57&280 (10\%)&200 (7.123\%); 400 (14.245\%)&(10-20) (25-8) (50-4) (10-40) (25-16) (50-8)&\\
		Spam$\_20^3$&3,160&57&632 (20\%)&200 (6.329\%); 400 (12.658\%)&(10-20) (25-8) (50-4) (10-40) (25-16) (50-8)&\\
		Spam$\_40^3$&4,207&57&1,679 (40\%)&200 (4.754\%); 400 (9.508\%)&(10-20) (25-8) (50-4) (10-40) (25-16) (50-8)&\\
		\cline{1-6}
		Australian	&690&15&\multicolumn{1}{|c||}{-}&30 (4.347\%); 60 (8.696\%)&(1-30) (5-6) (10-3) (1-60) (5-12) (10-6)&\\
		\cline{1-6}
		Ionosphere	    &351&33& \multicolumn{1}{|c||}{-}&20 (5.698\%); 40 (11.396\%)&(1-20) (5-4) (10-2) (1-40) (5-8) (10-4)&\\
		\cline{1-6}
		Iris			&150&4&\multicolumn{1}{|c||}{-}&10 (6.667\%); 20  (13.333\%)&(1-10) (5-2) (10-1) (1-20) (5-4) (10-2)& \\
		\cline{1-6}
		Isolet5			&1,559&618&\multicolumn{1}{|c||}{-}&75 (4.810\%); 150  (9.622\%)&(3-25) (5-15) (25-3) (5-30) (10-15) (25-6)& \\
		\cline{1-6}
		\cline{1-6}
		Synth. Distrib. 1& 1,000; 5,000; 10,000& 2 &  1\%; 5\%; 10\%&5\%; 10\%&\multicolumn{1}{|c|}{-}&\\
		Synth. Distrib. 2&1,000; 5,000; 10,000& 2 &  1\%; 5\%; 10\%&5\%; 10\%&\multicolumn{1}{|c|}{-}& \\
		\hline
	\end{tabular}
	\label{tab:datasets}
\end{table*}
\end{landscape}

\noindent{\bf Outlier detection methods.} We considered seven well-known outlier detection methods, classified into four categories: \\
{\em (1) Statistical methods}: 3$\sigma$ where observations outside the mean and $\pm 3$ standard deviation are considered outliers;  Boxplot with inner ($Q1 - 1.5\times IQR$) and outer ($Q3 + 1.5\times IQR$) fences \cite{tukey}, and  Chi-Square ($\chi^2$) defined in \cite{grubbs}; \\
{\em (2) Deviation-based methods}: MAD using the median absolute deviation as defined in  \cite{Carling};\\
{\em (3) Distance-based methods}: Mahalanobis distance which estimates how far each observation is from the center (centroid in multivariate space) \cite{Mahalanobis}, and  K-Means \cite{Hartigan}; \\
{\em (4) Density-based methods}: LOF defined in \cite{Breunig2000}. \\
We use 10\% of the data size as the threshold for the number of outliers to be detected by Mahalanobis, LOF and K-Means.
%

\noindent{\bf Sampling.} We study three types of sampling schemes:
{\em (1) Random sampling} with 100 random samples with real-world sample sizes shown in  Table \ref{tab:datasets} and for synthetic data, 1\%, 5\%, and 10\% of the data size;
{\em (2) Block sampling} with 100 block samples and six parameter pairs of block number and size (see Table \ref{tab:datasets});
{\em (3) Partitioning} where each dataset is divided into 5, 10, and 20 subsets.

\noindent{\bf Parameters:} We  vary $psize$, the sample size percentage of the whole dataset size, and parameter settings of each sampling scheme. For the synthetic dataset, we  vary the distributions and the rate of true outliers to understand how the resilience changes with an increasing number of true outliers.  

\subsection{Summary of Experimental Results:}  
In Section~\ref{section:resilienceVsF1}, we show that resilience to sampling is complementary to traditional metrics such as precision and recall. For example, some methods have high precision and/or recall when trained on the entire dataset but have poor performance when trained over a sample.

In Section~\ref{section:variety}, we note that resilience to sampling varies with method, dataset, sampling scheme and sample size. Most methods are not resilient to sampling thereby opening a promising research area of designing resilient algorithms.

In Section~\ref{section:size}, we show how sampling strategy and sample size affect resilience. Random sampling is often preferable to block sampling or partitioning. As expected, increasing sample size often improved resilience. Most algorithms are not resilient for small samples. Section \ref{section:true} shows that our resilience estimator is accurate and has low variance even for small samples.
Section \ref{section:em-based} shows that our two-coin idea for outlier ensemble using crowdsourcing is promising and provides good resilience estimates for  ensemble performance.


\begin{figure*}[!t]
	\centering
	\includegraphics[width=1.05\linewidth]{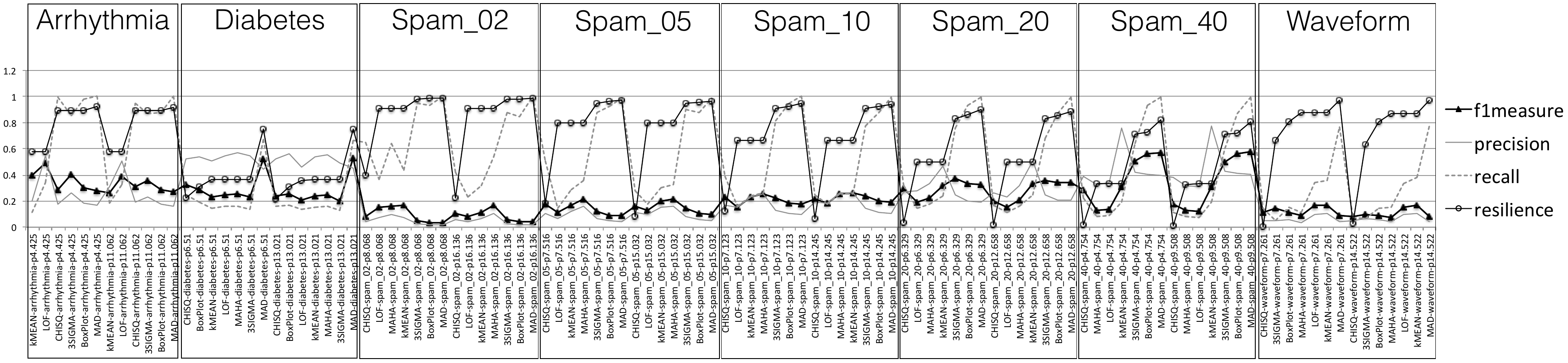}
	\caption{Resilience wrt random sampling and quality metrics  for real-world datasets with available Ground Truth}\label{fig:quality-metrics-vs-true-resilience}
\end{figure*} 

\begin{figure}[!t]
	\centering
	\includegraphics[width=\linewidth]{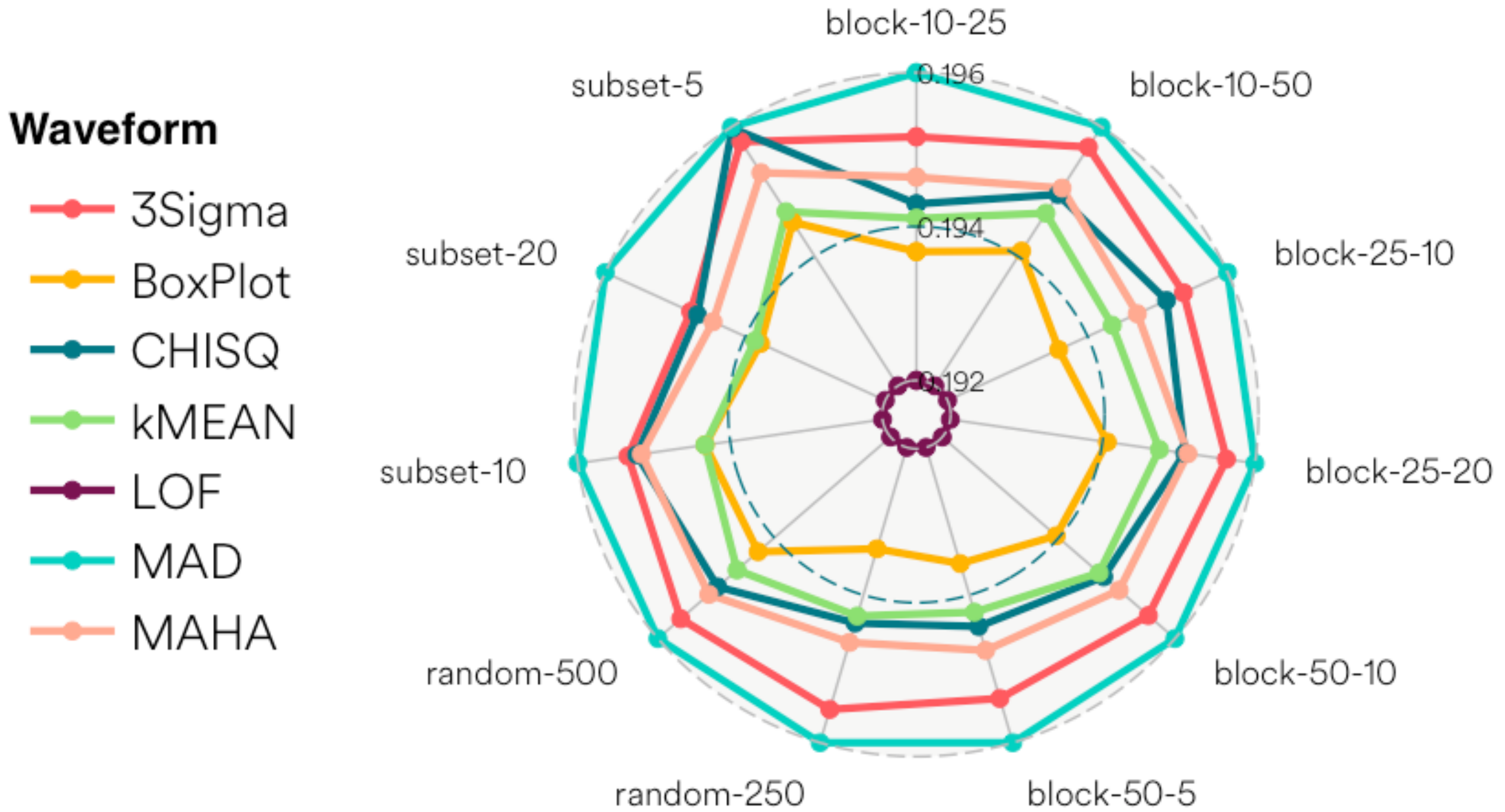}
	\includegraphics[width=\linewidth]{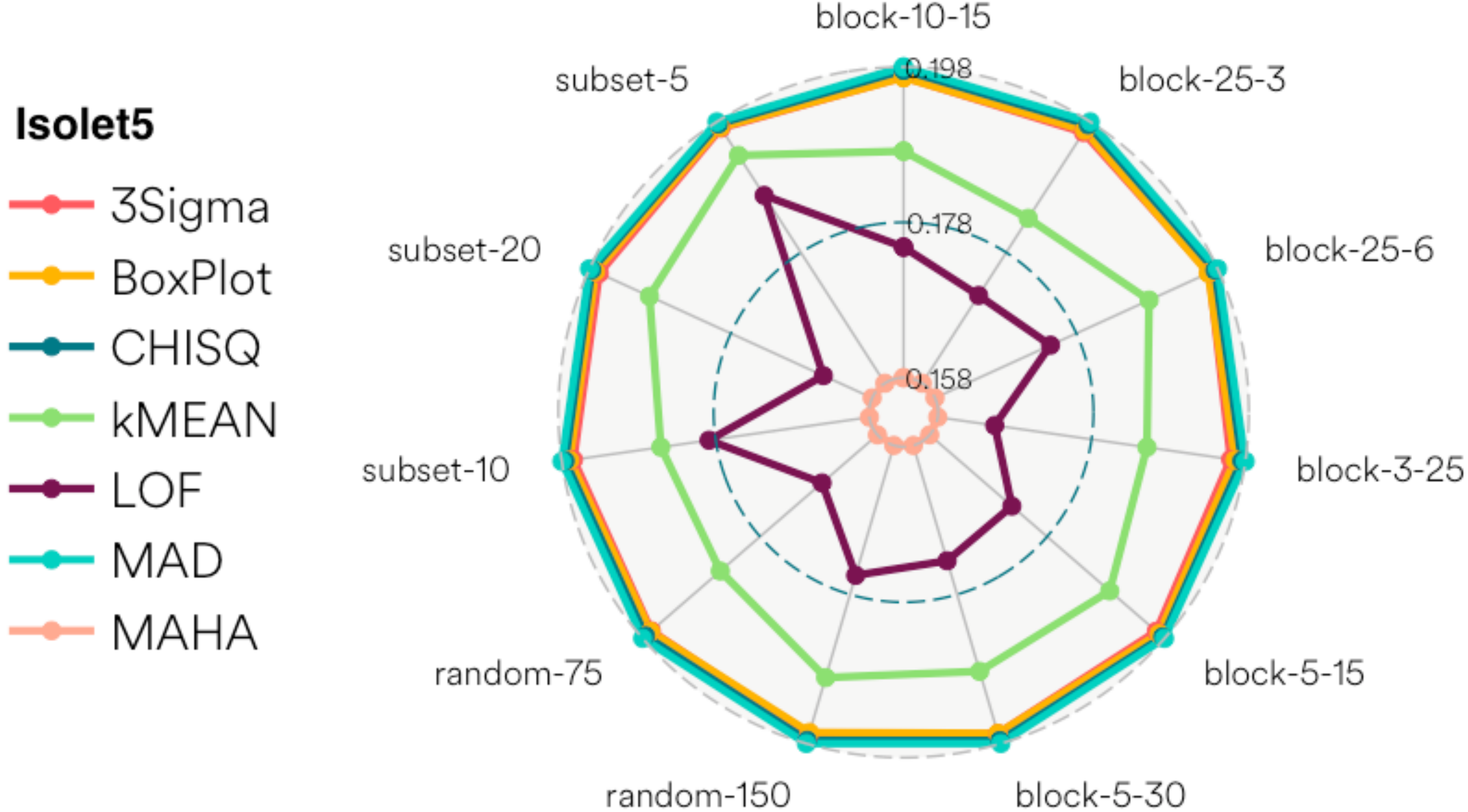}
	\caption{Resilience estimates for Waveform and Isolet5 datasets.  Univariate  statistical methods consistently have higher resilience compared to multivariate methods. LOF and MAHA are least resilient for Waveform and Isolet5 respectively). 
	}\label{fig:waveform}
\end{figure}

\subsection{Resilience is complementary to quality metrics} \label{section:resilienceVsF1}
We used the real-world datasets with available ground truth and computed classical performance metrics (F1-measure, precision and recall) as well as resilience calculated with Eq. (\ref{rho}) for different sample sizes. 
We find that outlier methods with relatively low F1-measure, precision and recall  can have high resilience (e.g.\ univariate methods in Spam dataset) and vice-versa (e.g., CHISQ in Waveform and Spam datasets) (Fig. \ref{fig:quality-metrics-vs-true-resilience}).
Increasing the prevalence of outliers and data size logically increases the quality metrics but not necessarily the resilience (e.g., resilience decreased from Spam\_02 to Spam\_40). This suggests a need for better understanding of resilience and its estimation when ground truth is unavailable.

\subsection{Resilience of Outlier Methods}\label{section:variety}
Here, we consider only the real-world datasets without ground truth and computed the resilience estimates for each method using 100 samples from each sampling strategy.

\begin{figure*}[!t]
\begin{tabular}{ccc}
	\includegraphics[width=.33\linewidth]{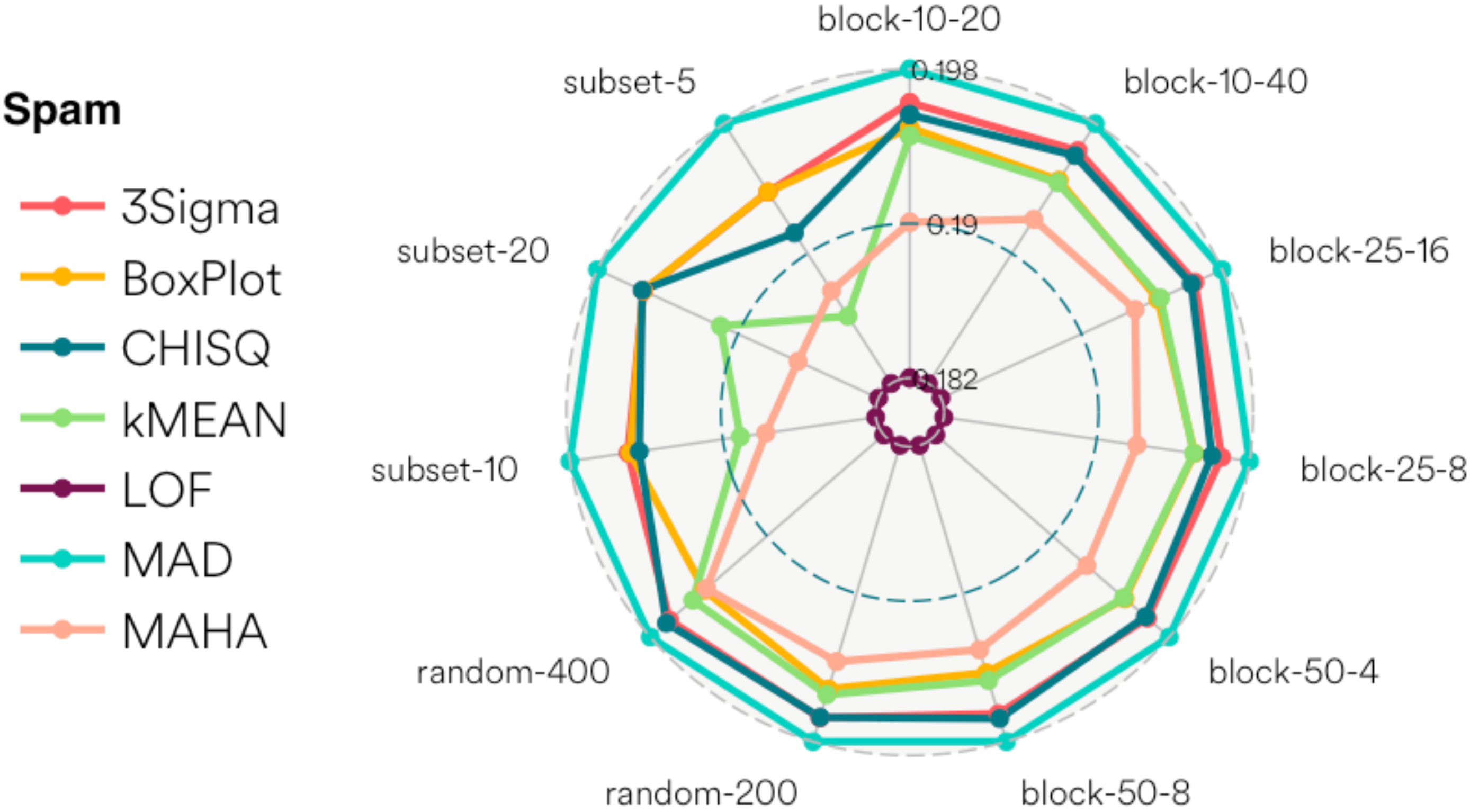} &
	\includegraphics[width=.33\linewidth]{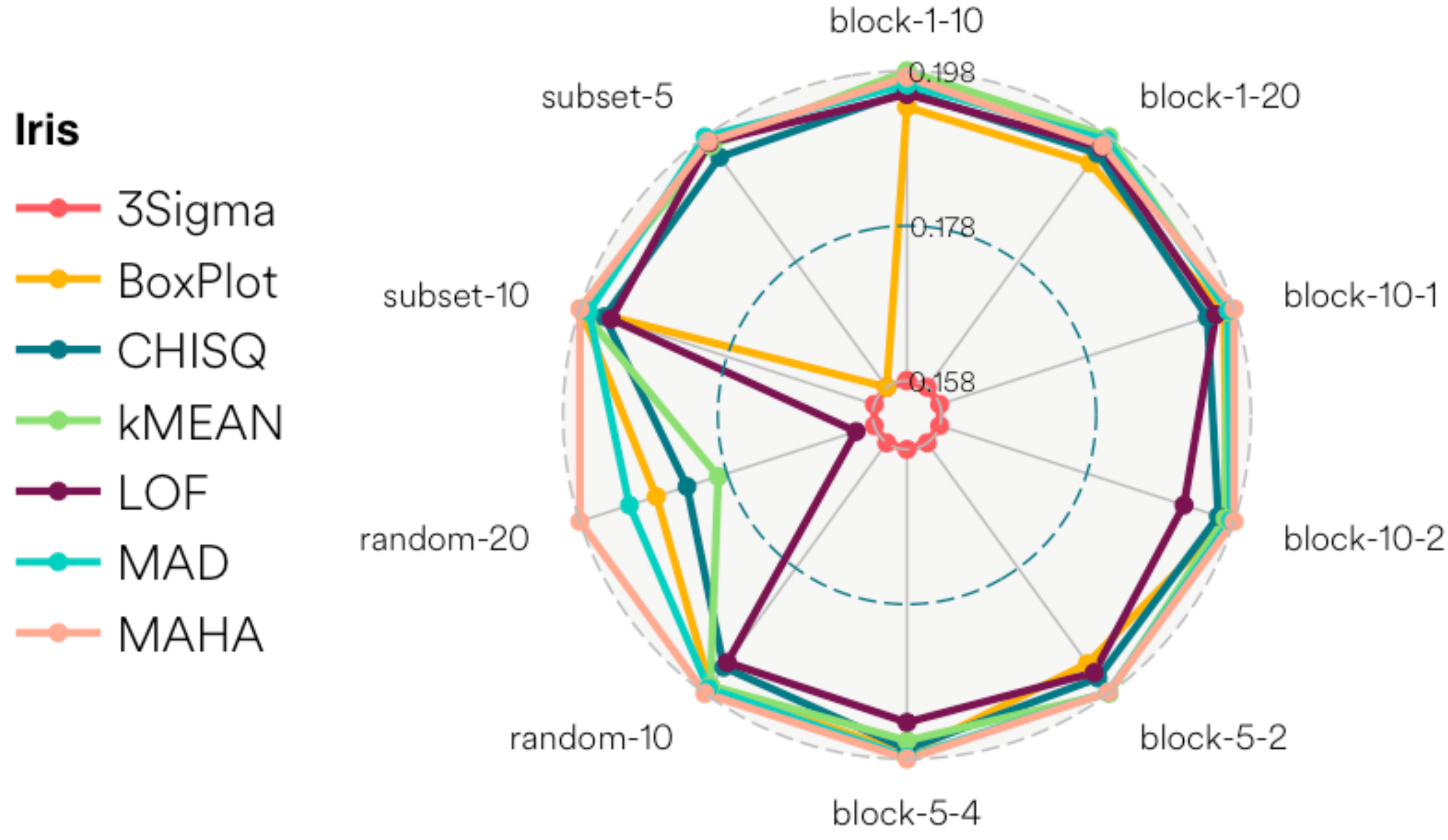} &
	\includegraphics[width=.33\linewidth]{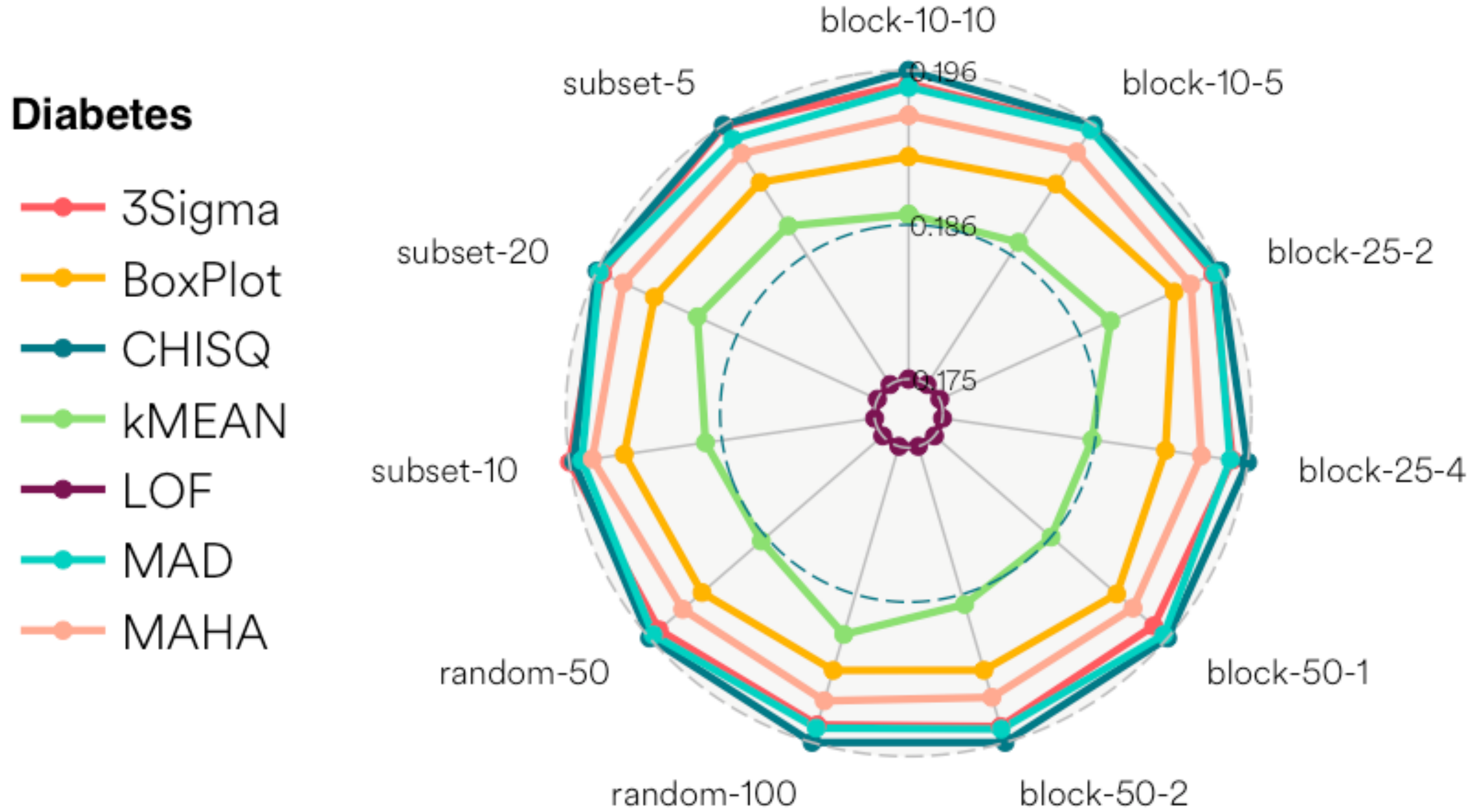} \\
	\includegraphics[width=.33\linewidth]{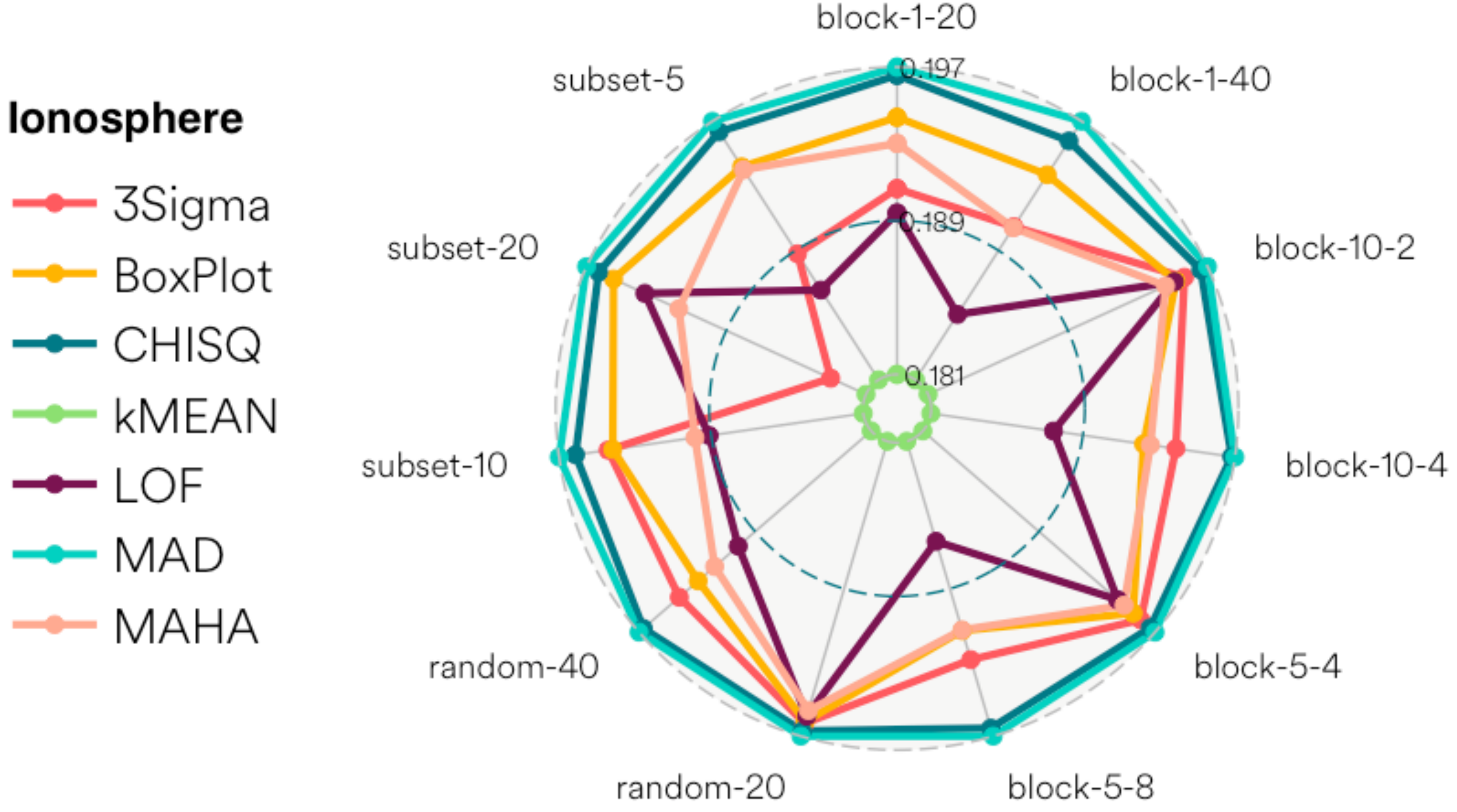}&
	\includegraphics[width=.33\linewidth]{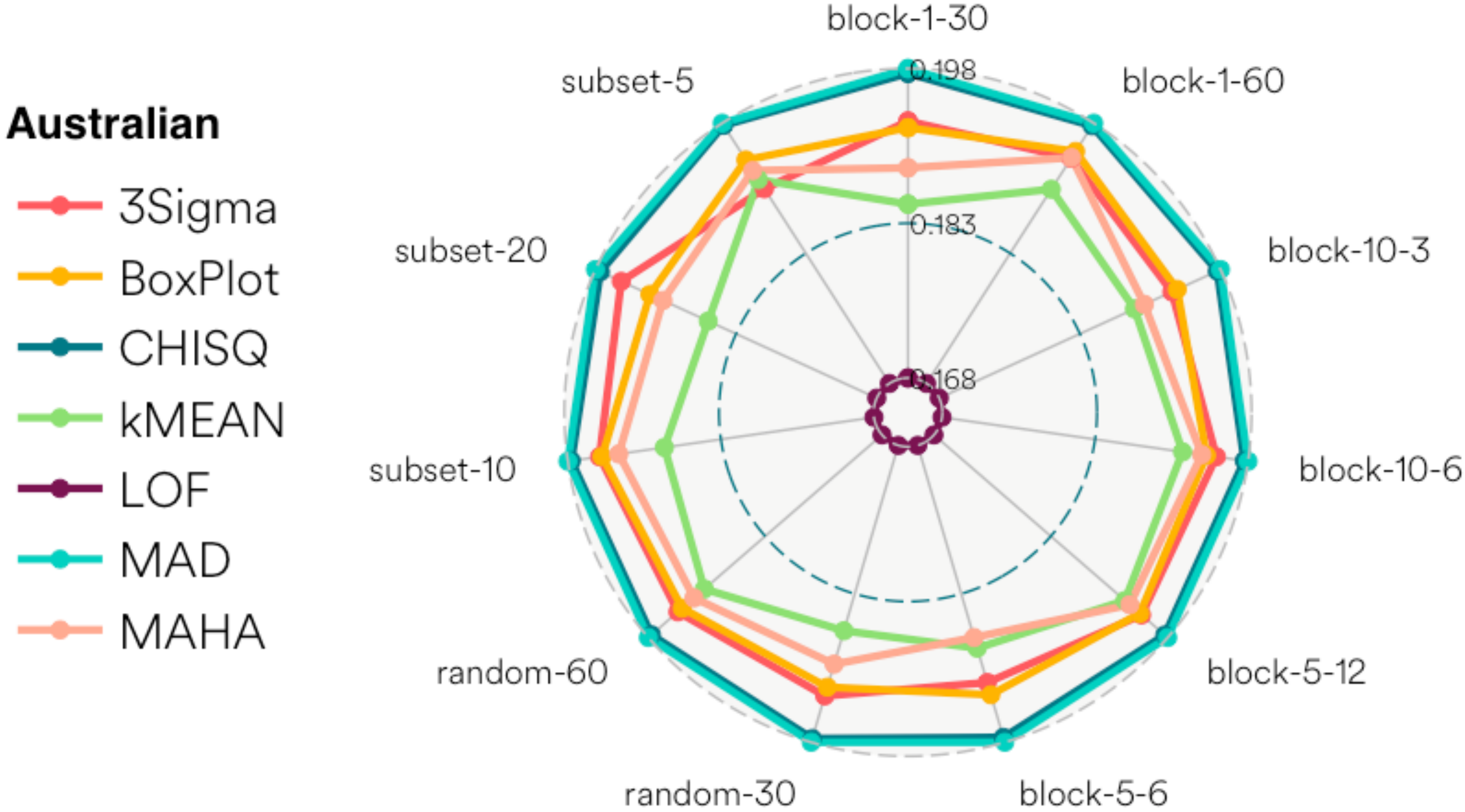}&
	\includegraphics[width=.33\linewidth]{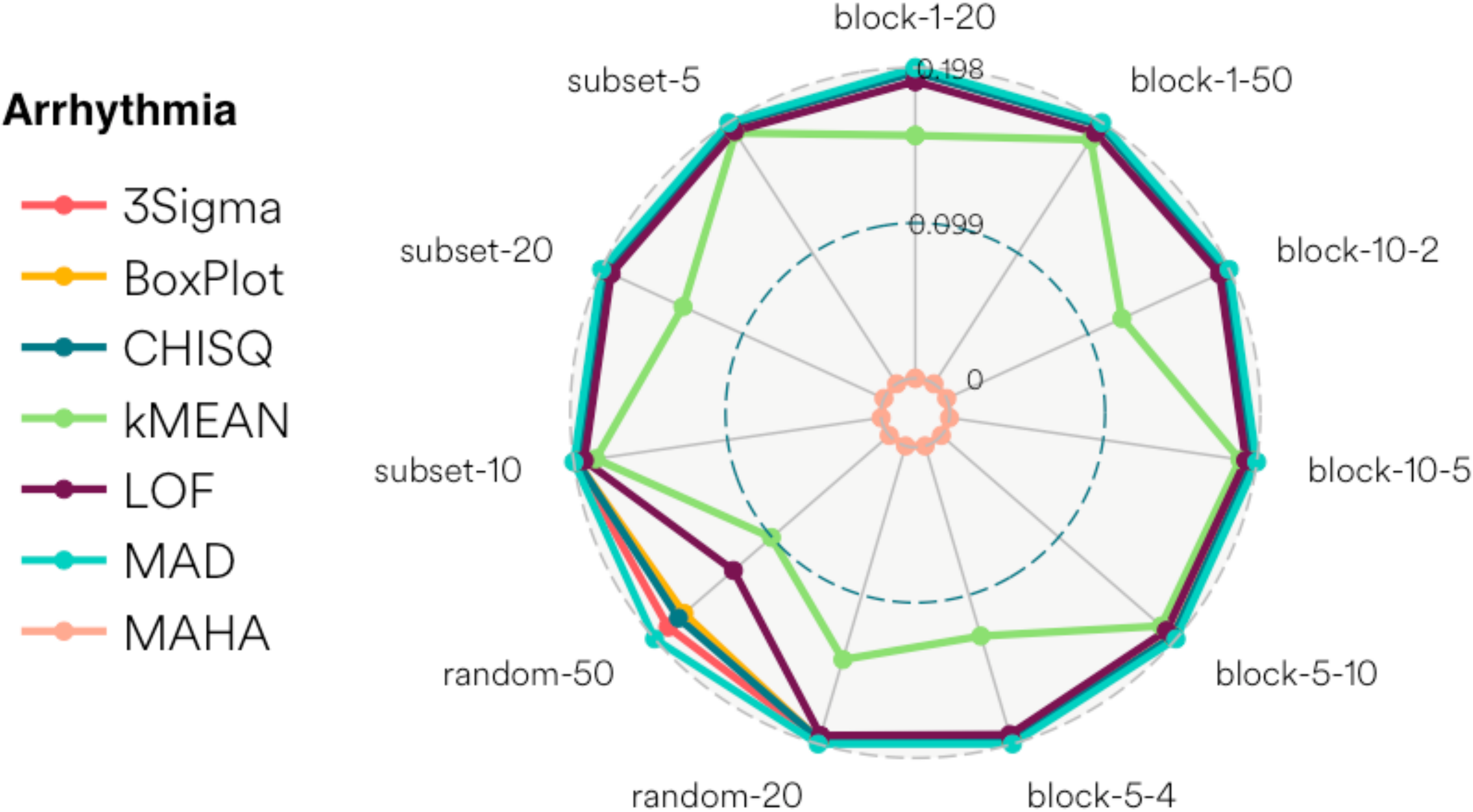}
\end{tabular}
	\caption{Resilience estimates of outlier detection methods wrt sampling schemes and sample sizes for six real-world datasets.}
	\label{fig:RW-allmethods} 
\end{figure*}

Figure \ref{fig:waveform} shows the results for the Waveform and Isolet5 datasets, and Figure \ref{fig:RW-allmethods} for the other datasets. 
We find that resilience to sampling varies greatly depending on sample size and data and outlier distributions. In the figures, resilience estimates 
are represented as the distance from the center of the plot. Each axis represents a particular sampling strategy and sample size; a colored line in the plot shows the trend of the resilience estimate when both the sampling scheme and sample size change. The area defined by each colored line represents the resilience estimate of a method relative to the dataset and sampling scheme: the smaller the area, the less resilient. 
Focusing on the Waveform and Isolet5 datasets, MAD has the highest resilience estimaates for all sampling strategies. LOF and MAHA had the lowest resilience for Waveform and Isolet5 respectively. Particularly for Isolet5, statistical and univariate methods such as MAD and $3\sigma$ seem to be more resilient to random and block sampling.
The resilience of  BoxPlot, distance- and density-based multivariate methods seem to be  affected by the data characteristics:  Isolet5 has wide skewness (asymmetry) and kurtosis (shape) ranges of ([-2.22;31.05],[1.44;1220.16]). This affected MAHA which is less resilient than LOF in Isolet5.  On the other hand, in Waveform which is more symmetric and flat, with ([-0.24;0.28],[1.49;3.11]) for (skewness, kurtosis) ranges, MAHA is more resilient than LOF.  For each dataset, the ranking of the methods by resilience is generally preserved across sampling strategies. 
The resilience estimates tend to slightly increase with sample size for all sampling schemes but at different rates. This suggests that the sampling scheme and sample size may have different impacts on a method's resilience.

\subsection{Impact of sample scheme and sample size}\label{section:size}
\begin{figure*}[!t]
	\raggedleft	\includegraphics[width=.99\linewidth]{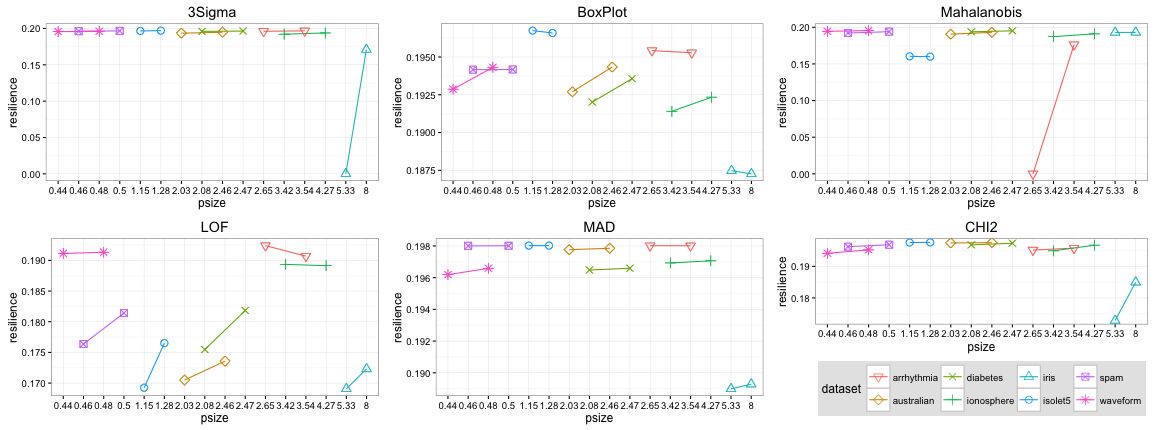}
	\caption{Impact of random sampling and sample sizes (as percentage of the whole dataset size) on the resilience estimates of the methods applied to the real-world datasets.}
	\label{fig:random-resilience-size} 
\end{figure*}

\begin{figure*}[!t]
	\raggedleft	\includegraphics[width=.99\linewidth]{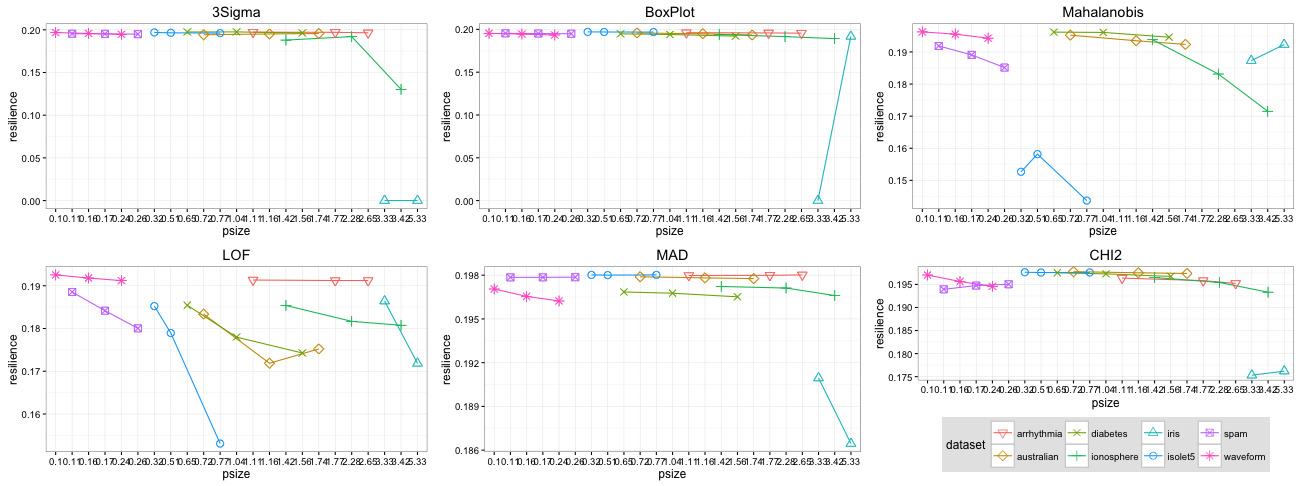}
	\caption{Impact of partitioning and partition sizes (as percentage of the whole dataset size) on the resilience estimates of the methods applied to the real-world datasets.}
	\label{fig:subset-resilience-size} 
\end{figure*}

{\bf  Random sampling}. From Fig.\ \ref{fig:random-resilience-size},  we find that mean resilience estimates increase with sample size. While all methods have relatively low resilience estimates, the $3\sigma$ method has the highest, consistently across datasets, sampling stratgegies and sample size percentage. $\chi^2$ and MAD have similarly high resilience estimates but are more dataset-dependent. LOF has the lowest  estimates and is very dataset-dependent, with the greatest variability. This may be because LOF relies on local density properties which are lost in the sampling process.\\
{\bf  Block sampling}.
LOF resilience estimates are again data dependent and impacted by block number and sample size, with no apparent trend (not shown). 
$3\sigma$ still attains the highest resilience (except for Iris) and is stable across datasets, block number and sample size. The resilience estimates for MAHA, $\chi^22$, MAD and BoxPlot are more variable and data dependent. \\
{\bf  Partitioning}. We used 5, 10 and 20 subsets, correspondingly reducing the size of each subset. The trend in Fig.\ \ref{fig:subset-resilience-size} is reversed and is more apparent for the multivariate methods (LOF and MAHA) for which intrinsic properties such as distance and density are not preserved in the samples: their resilience estimates decrease when the  size of the subsets decreases; the resilience ranges of $3\sigma$, $\chi^2$, MAD, and BoxPlot are again  compact and decreases slightly when the number of subsets decreases (and their sizes increase).

%
%

\begin{figure*}[!t]
	\centering
	\includegraphics[width=.8\linewidth]{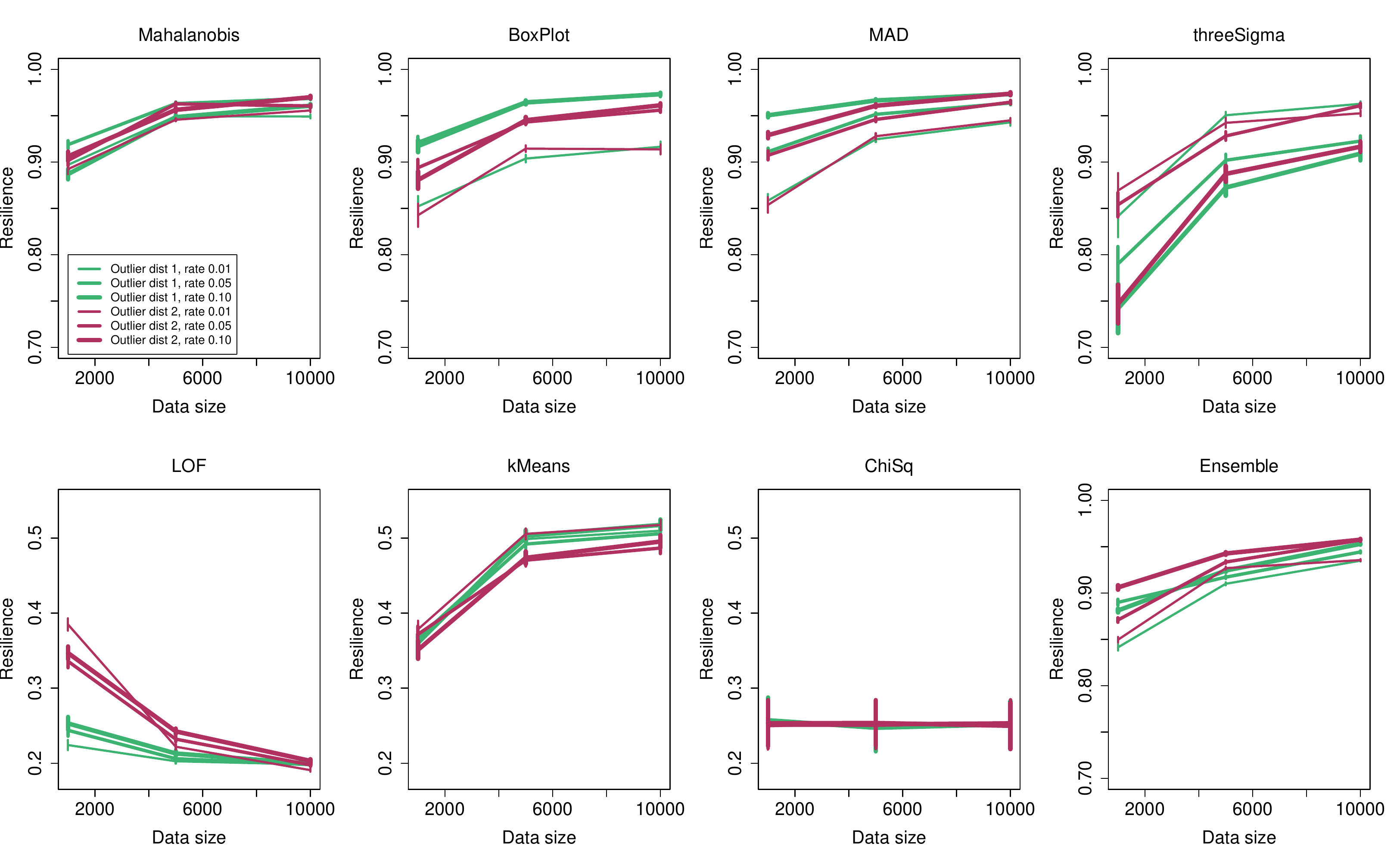}
	\caption{Mean and standard deviation of resilience estimates for each method computed from five subsets (1/5 of the full dataset size) over 100 simulations for two outlier distributions and three rates of outlier injection.}\label{fig:resilience-synthetic}
\end{figure*}

The key observation is that statistical methods tend to be more resilient than multivariate density- and distance-based methods to random and block sampling.
Furthermore, $3\sigma$ has the highest resilience estimates and is not much affected by the data characteristics. Partitioning and block sampling work better for BoxPlot than random sampling. Block sampling with LOF should be avoided mainly due to the loss of the  local density properties in the samples.

\subsection{Accuracy of Resilience Estimates} \label{section:true}

{\bf  Accuracy wrt true outlier distribution and prevalence.} 
We controlled the rate of true outliers  injected in the synthetic datasets and, for each method, computed the resilience and checked its accuracy. Fig. \ref{fig:resilience-synthetic} shows resilience estimates for the two outlier distributions (same color scheme as Fig. 3) and three outlier rates (increasing line width) for the case of partitioning. Five subsets were used. The MAHA resilience seems to be high for all our settings of outlier distribution and prevalence rates. Using subsets for detection did not seem to affect this. To a lesser extent, this is also true for the MAD and BoxPlot methods as well, although there is some difference between the two outlier distributions. Their high resilience could be due to the symmetry and shape of the synthetic data distribution that are somehow preserved in the samples. This confirms our previous findings from real-world datasets. Both LOF and the $\chi^2$  methods seem to have lower detection power in our study, and exhibit noticeable low resilience when using outlier detection from subsets (in particular LOF for Distribution 1). This is mainly due to the fact that LOF is a density-based outlier detection method and the density property is somehow lost in the samples.  Both the LOF and $3\sigma$ methods show large variation in resilience depending on the outlier prevalence rate,  decreasing  as the rate increases.  With increasing full dataset size, LOF resilience significantly decreases in opposition with the other methods. This suggests a direction for future work to include density-based sampling such as that in \cite{Ros2016349} to fairly test the resilience of LOF. More importantly, it shows that a joint choice of outlier detection method with appropriate sampling scheme is crucial.

{\bf  Accuracy wrt to the number of samples.} Tables III and IV show mean square errors (MSE) of resilience estimates for $3\sigma$ and BoxPlot using synthetic and real-world datasets. The low MSEs demonstrate that our model for computing resilience estimates is robust and has very good accuracy. Increasing the number of samples does not necessarily increase the accuracy of the estimates.  Increasing the number of samples did not necessarily increase the accuracy of the estimates. 

 \begin{table}[ht]
\centering 
\small
\begin{tabular}{cclcccc}
  \hline
Outlier& Outlier & Data  & \multicolumn{2}{c}{10 samples} & \multicolumn{2}{c}{20 samples} \\
distribution & rate & size & $3\sigma$ & Boxplot & $3\sigma$ & Boxplot\\ 
  \hline
 1 & 0.05  & 1000 & 1.2e-02 & 1.1e-03 & 1.8e-02 & 3.8e-03 \\ 
& & 10,000  & 3.8e-04 & 3.5e-05 & 7.1e-04 & 6.9e-05 \\ 
& 0.1  &  1000 & 2.8e-02 & 1.6e-03 & 2.0e-02 & 5.2e-03 \\ 
& & 10,000 & 2.1e-03 & 2.1e-05 & 5.0e-03 & 3.5e-05 \\ \hline
 2 & 0.05 & 1000 & 5.7e-03 & 1.8e-03 & 1.0e-02 & 5.9e-03 \\ 
& & 10,000 & 1.4e-04 & 8.7e-05 & 2.8e-04 & 1.4e-04 \\ 
& 0.1  & 1000 & 1.1e-02 & 9.1e-04 & 1.8e-02 & 2.8e-03 \\ 
& & 10,000 & 4.0e-04 & 4.9e-05 & 6.5e-04 & 7.9e-05 \\ 
   \hline
\end{tabular}\label{tab:MSEsynthetic}
\caption{MSE of resilience estimates for $3\sigma$ and Boxplot over multiple synthetic subset samples.}
\end{table}

\begin{table}[ht]
\centering
\small
\begin{tabular}{lcccccc}
  \hline
& \multicolumn{2}{c}{5 samples} & \multicolumn{2}{c}{10 samples} & \multicolumn{2}{c}{20 samples}  \\
 & $3\sigma$ & Boxplot & $3\sigma$ & Boxplot & $3\sigma$ & Boxplot \\  \hline
arrhythmia & 1.3e-04 & 3.2e-05 & 1.9e-04 & 9.9e-05 & 3.7e-04 & 2.4e-04 \\ 
  australian & 6.0e-03 & 1.0e-03 & 6.8e-03 & 1.7e-03 & 9.1e-03 & 3.3e-03 \\ 
  diabetes & 4.0e-04 & 9.3e-04 & 7.3e-04 & 1.9e-03 & 1.9e-03 & 5.8e-03 \\ 
  ionosphere & 1.2e-02 & 7.7e-04 & 2.5e-02 & 1.9e-03 & 3.6e-02 & 4.8e-03 \\ 
  iris & 6.4e-02 & 1.9e-02 &     NA & 7.7e-03 &     NA & 1.7e-02 \\ 
  isolet5 & 4.0e-05 & 8.4e-06 & 5.5e-05 & 1.0e-05 & 8.1e-05 & 2.2e-05 \\ 
  spam & 2.0e-05 & 1.1e-07 & 2.3e-05 & 2.0e-07 & 5.1e-05 & 6.1e-07 \\ 
  waveform & 3.5e-04 & 2.9e-04 & 5.6e-04 & 4.0e-04 & 9.5e-04 & 1.1e-03 \\ 
   \hline
\end{tabular}\label{tab:MSEreal}
\caption{MSE of resilience estimates for $3\sigma$ and Boxplot over multiple real-world subset samples.}
\end{table}

\begin{figure}
	\centering
	\includegraphics[width=0.7\linewidth]{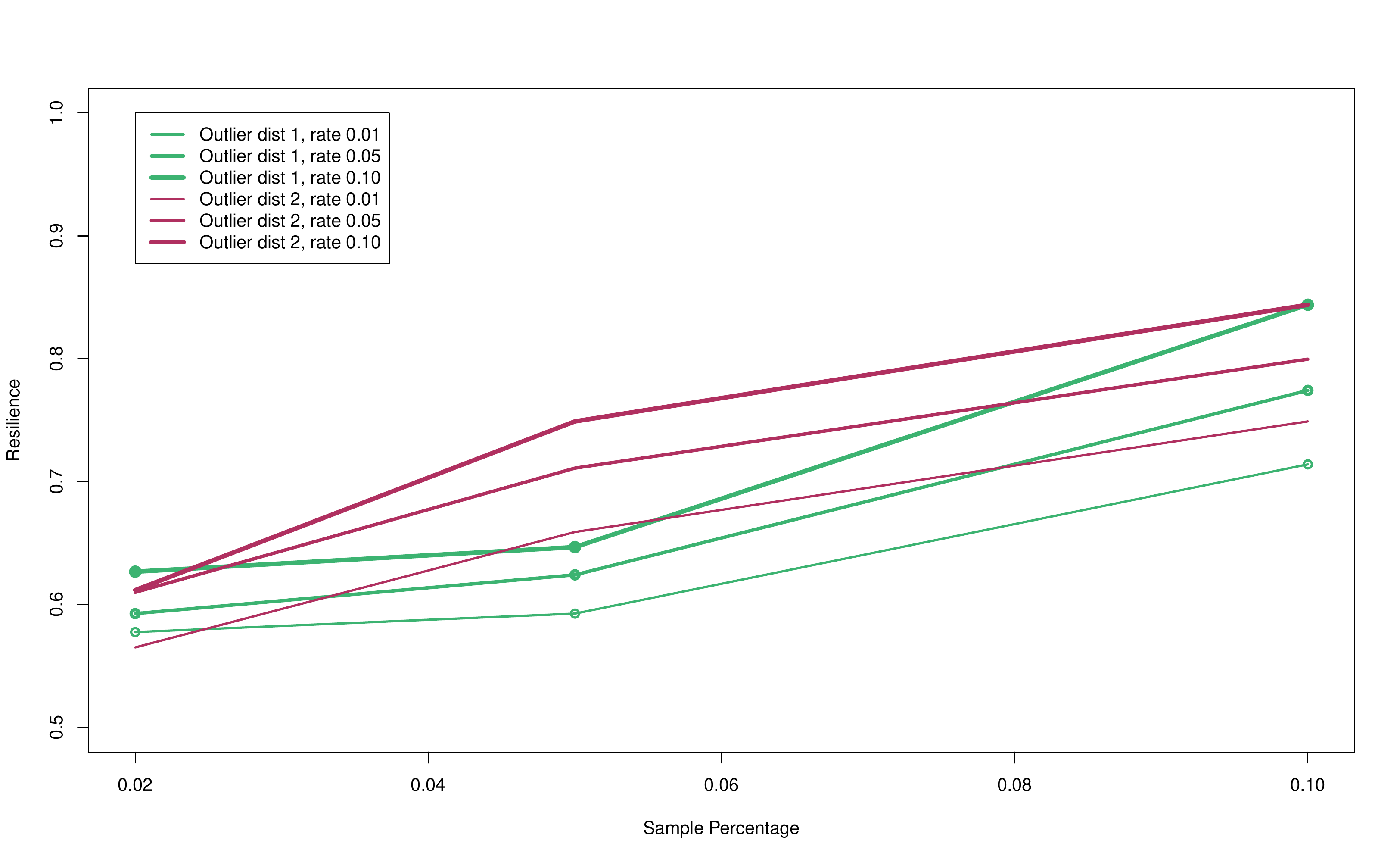}
	\caption{Outlier Ensemble resilience for Random Sampling}
	\label{fig:resilienceEnsembleRandomSample}
\end{figure}

\subsection{Evaluation of EM-Based Ensembling}\label{section:em-based}
With the synthetic data, we evaluate our EM-based ensembling approach on two aspects:
(a) its resilience compared to the component methods and 
(b) its accuracy at estimating the sensitivity and specificity.
We varied data sizes (1,000, 5,000 and 10,000), outlier prevalence rate (1\%, 5\%, and 10\%) and considered two sampling schemes:
(a) random sampling with sampling fraction of 1\%, 5\%, and 10\% and
(b) partitioning with the data split into 5 and 10 equi-sized subsets.
For each scenario, we used 100 different samples and averaged the results.
While we only show the results for two representative distributions from Fig.~\ref{fig:syn-data}, the results were similar for the others.

{\bf Resilience of Outlier Ensembles:}
Fig.~\ref{fig:resilienceEnsembleRandomSample} for random sampling
and Fig.~\ref{fig:resilience-synthetic} (bottom right) for partitioning show high resilience compared to its component algorithms. As expected, increasing the sample size increases the resilience, an increase that quickly plateaus.  This shows that it is feasible to build resilient outlier ensembles using only a fraction of the data size.
We also evaluated how the composition of the outlier ensembles affects the resilience by randomly choosing different subsets of outlier detection methods. We find that regardless of the mix of resilient and less resilient methods, the outlier ensemble retains its high resilience providing the best of both worlds in terms of both accuracy and resilience.

{\bf Accuracy of Resilience Estimation:}
Tables~\ref{tbl:emBasedExpResultsRandomSampling} and \ref{tbl:emBasedExpResultsSubsetting} show the RMSE of sensitivity and specificity estimates of the EM-based ensemble for random and subset sampling under different outlier distributions. The key observation is that the RMSE is very small for all settings suggesting that the EM-based algorithm can estimate the ground truth  accurately from the ensembles.

\begin{landscape}
\begin{table*}[!t]
\centering
	\caption{EM-based ensembling for Random Sampling:  RMSE of sensitivity and specificity estimates of the ensemble.
	}
	\small
	\begin{tabular}{|c|c|c|c|c|c|c|c|}
		\cline{3-8}
		\multicolumn{1}{c}{} & 	\multicolumn{1}{c}{} & 	\multicolumn{6}{|c|}{RMSE($\alpha$), RMSE($\beta$)} \\
		\hline
		Outlier Distribution & $N$ & $s=5\%,pt=1\%$ & $s=5\%,pt=5\%$ & $s=5\%,pt=10\%$ & $s=10\%,pt=1\%$ & $s=10\%,pt=5\%$ & $s=10\%,pt=10\%$ \\ \hline
		Distribution 1 & 1000 & 0.01, 0.03 & 0.01, 0.02 & 0.01, 0.02 & 0.01, 0.02 & 0.01, 0.02 & 0.01, 0.02 \\ \hline
		Distribution 1 & 10000 & 0.01, 0.03 & 0.01, 0.02 & 0.01, 0.02 & 0.01, 0.02 & 0.01, 0.02 & 0.01, 0.02 \\ \hline
		Distribution 2 & 1000 & 0.01, 0.03 & 0.01, 0.02 & 0.01, 0.02 & 0.01, 0.02 & 0.01, 0.02 & 0.01, 0.02 \\ \hline
		Distribution 2 & 10000 & 0.01, 0.02 & 0.01, 0.02 & 0.01, 0.02 & 0.01, 0.02 & 0.01, 0.02 & 0.01, 0.02 \\ \hline
	\end{tabular}
	\label{tbl:emBasedExpResultsRandomSampling}
\end{table*}
\begin{table*}[!t]
\centering
	\caption{EM-based ensembling for Subset Sampling:  RMSE of sensitivity and specificity estimates of the ensemble. 
	}
	\small
	\begin{tabular}{|c|c|c|c|c|c|c|c|}
		\cline{3-8}
		\multicolumn{1}{c}{} & 	\multicolumn{1}{c}{} & 	\multicolumn{6}{|c|}{RMSE($\alpha$), RMSE($\beta$)} \\
		\hline
		
		Outlier Distribution & $N$ & $s=5,pt=1\%$ & $s=5,pt=5\%$ & $s=5,pt=10\%$ & $s=10,pt=1\%$ & $s=10,pt=5\%$ & $s=10,pt=10\%$ \\ \hline
		Distribution 1 & 1000 & 0.06, 0.04 & 0.04, 0.06 & 0.04, 0.05 & 0.06, 0.04 & 0.04, 0.06 & 0.04, 0.05 \\ \hline
		Distribution 1 & 5000 & 0.06, 0.04 & 0.04, 0.05 & 0.04, 0.05 & 0.06, 0.04 & 0.04, 0.05 & 0.04, 0.05 \\ \hline 
		Distribution 1 & 10000 & 0.06, 0.04 & 0.04, 0.06 & 0.03, 0.05 & 0.06, 0.04 & 0.04, 0.06 & 0.03, 0.05 \\ \hline
		Distribution 2 & 1000 & 0.06, 0.04 & 0.04, 0.06 & 0.01, 0.07 & 0.06, 0.04 & 0.04, 0.06 & 0.01, 0.07 \\ \hline
		Distribution 2 & 5000 & 0.06, 0.04 & 0.05, 0.06 & 0.01, 0.07 & 0.06, 0.04 & 0.05, 0.06 & 0.01, 0.07 \\ \hline
		Distribution 2 & 10000 & 0.06, 0.04 & 0.05, 0.06 & 0.01, 0.07 & 0.06, 0.04 & 0.05, 0.06 & 0.01, 0.07 \\ \hline
	\end{tabular}
	\label{tbl:emBasedExpResultsSubsetting}
\end{table*}
\end{landscape}

\section{Related work}
\label{sec:relWork}

The outlier detection literature is vast and this section is not exhaustive.
A comprehensive survey can be found in \cite{Chandola:2009} and \cite{aggarwal2015outlier} and a recent comparative study in \cite{DAMI16}. 
Prior research can be broadly categorized into three types. 
The parametric approach builds statistical models based on some assumptions about the dataset. While conceptually simple, they suffer from shortcomings such as 
simplifying assumptions about data distributions, less emphasis on interpretability and poor scalability.
An alternate approach involves distance-based techniques that do not make any prior assumption on data distribution. 
However, this approach has scalability issues as it might require computation of distances between all pairs of points.
There has been slew of techniques to scale distance-based methods including indices (e.g.\ \cite{ramaswamy2000efficient}), pruning (e.g.\ \cite{bay2003mining}) and approximations. See \cite{knorr2000distance} for a summary of such techniques.
Yet another approach is based on ``local'' outliers that considers the distances (or densities) of observations in its neighborhood.
LOF \cite{Breunig2000} and LOCI \cite{Papadimitriou2003a} are canonical examples of this approach. 
However, there is few prior work on sampling for speeding up outlier detection.
Most are focused on distance-based outlier detection techniques.
\cite{kollios2003efficient} showed that a biased sample where the sampling probability is proportional to the density of the dataset 
often works well for diverse data mining tasks such as outlier detection and clustering.
\cite{wu2006outlier} proposed a sampling technique that efficiently approximated the distance to the $k$-th nearest neighbor.  
\cite{sugiyama2013rapid} evaluated a number of distance-based outlier detection algorithms and found that sampling-based approaches 
outperformed many state-of-the art algorithms in both efficiency and effectiveness. 

Outlier ensembles is an emerging research topic with a number of challenging issues. 
One the earliest work \cite{lazarevic2005feature} adapted the idea of (feature) bagging to outliers by repeatedly sampling different sets of dimensions of the data for outlier detection. Some theoretical foundation for outlier ensembles is provided in \cite{zimek2014ensembles} and \cite{aggarwal2015theoretical}.
\cite{zimek2013subsampling} introduced the idea of subsampling to include diversity in the outlier ensembles and found that running an outlier ensemble on subsamples is more efficient than other sophisticated methods to obtain diverse ensembles.
It has been found that outlier detection on subsamples is even more efficient than running a single outlier detector on the entire dataset.
\cite{aggarwal2015theoretical} and \cite{zimek2013subsampling} also proposed interesting techniques for model and score aggregation for outlier detection.
However, none of the previous work characterized, quantified or estimated the resilience of outlier detection methods to a particular sampling technique.

	
	\balance
	\bibliographystyle{abbrv}
	\bibliography{biblio}

\end{document}